\documentclass[10pt,twocolumn,letterpaper]{article}

\usepackage{iccv}
\usepackage{times}
\usepackage{epsfig}
\usepackage{amsmath}
\usepackage{amssymb}
\usepackage{bm}
\usepackage{bbding}
\usepackage{multirow}
\usepackage{booktabs,graphicx}
\usepackage{animate}
\usepackage{tabu} 

\usepackage[pagebackref=true,breaklinks=true,letterpaper=true,colorlinks,bookmarks=false]{hyperref}

 \iccvfinalcopy 

\def\iccvPaperID{4514} 

\ificcvfinal\fi

\begin{document}
	
	\title{Self-supervised Learning to Bring Dual Reversed Rolling Shutter Images Alive}
	
	\author{Wei Shang$^{1,2}$, Dongwei Ren$^1$\thanks{Corresponding author: rendongweihit@gmail.com.} , Chaoyu Feng, Xiaotao Wang, Lei Lei, Wangmeng Zuo$^{1,3}$\\ 
		$^1$School of Computer Science and Technology, Harbin Institute of Technology\\   $^2$City University of Hong Kong \quad    $^3$Peng Cheng Laboratory, Shenzhen \\  
	}

\maketitle
\ificcvfinal\thispagestyle{empty}\fi

\begin{figure*}[!t]\footnotesize
	\setlength{\abovecaptionskip}{3pt} 
	\setlength{\belowcaptionskip}{0pt}
	\hspace{-1em}
	\begin{tabular}{l}
		\includegraphics[width=0.70\linewidth]{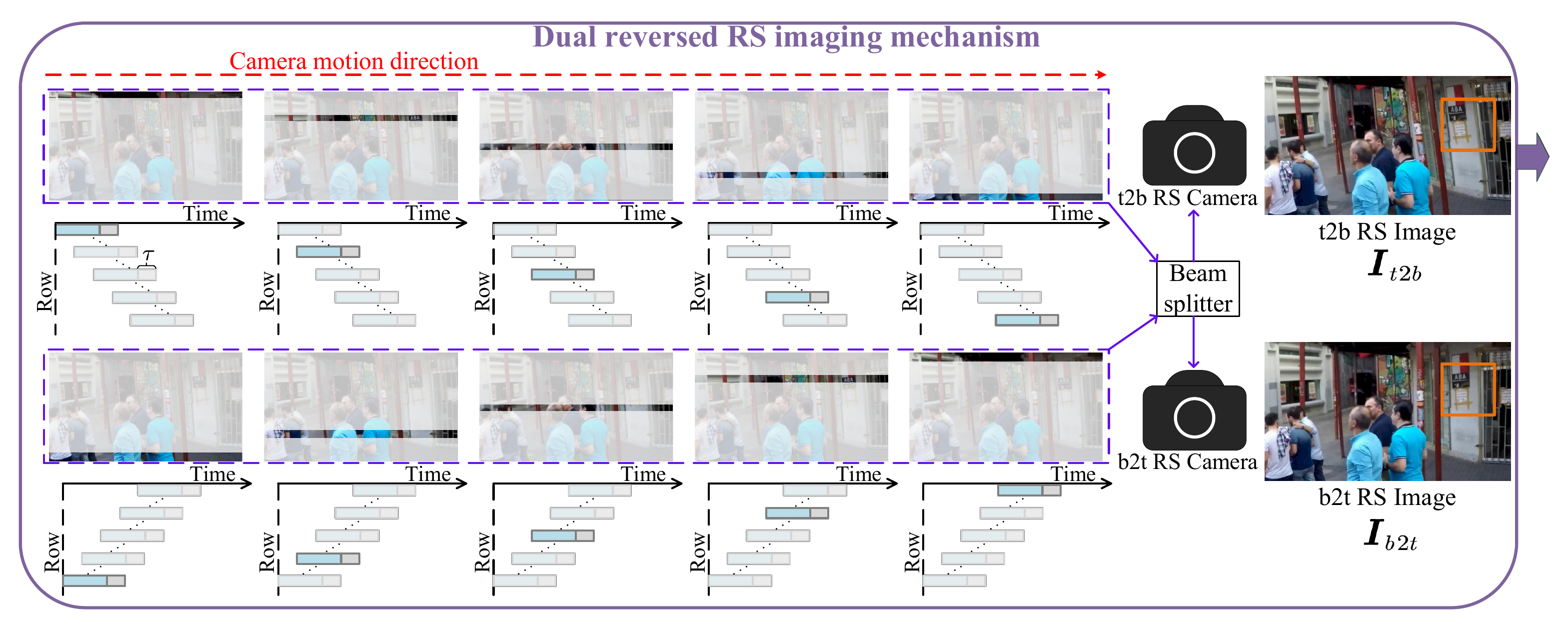}   
		\animategraphics[width=0.29\linewidth, autoplay, loop]{10}{fig/cropframework/}{1}{25}\\   
	\end{tabular}
	\caption{
		Illustration of capturing dual RS images with reversed scanning directions, \ie, top-to-bottom ($\bm{I}_{t2b}$) and bottom-to-top ($\bm{I}_{b2t}$). 
		In this work, we propose the first self-supervised learning method SelfDRSC to correct RS distortions.  
		In comparison to state-of-the-art supervised RS correction methods CVR \cite{Fan_2022_CVPR} and IFED \cite{zhong2022bringing}, our SelfDRSC can generate high framerate GS videos with finer textures and better temporary consistency. 
		The animated video results can be viewed in Adobe PDF reader. 
	}
	\label{fig:dual RS}
\end{figure*}
\begin{abstract}
	Modern consumer cameras usually employ the rolling shutter (RS) mechanism, where images are captured by scanning scenes row-by-row, yielding RS distortions for dynamic scenes. 
	%
	%
	To correct RS distortions, existing methods adopt a fully supervised learning manner, where high framerate global shutter (GS) images should be collected as ground-truth supervision. 
	%
	In this paper, we propose a Self-supervised learning framework for Dual reversed RS distortions Correction (SelfDRSC), where a DRSC network can be learned to generate a high framerate GS video only based on dual RS images with reversed distortions.   
	%
	In particular, a bidirectional distortion warping module is proposed for reconstructing dual reversed RS images, and then a self-supervised loss can be deployed to train DRSC network by enhancing the cycle consistency between input and reconstructed dual reversed RS images.  
	Besides start and end RS scanning time, GS images at arbitrary intermediate scanning time can also be supervised in SelfDRSC, thus enabling the learned DRSC network to generate a high framerate GS video. 
	Moreover, a simple yet effective self-distillation strategy is introduced in self-supervised loss for mitigating boundary artifacts in generated GS images. 
	%
	On synthetic dataset, SelfDRSC achieves better or comparable quantitative metrics in comparison to state-of-the-art methods trained in the full supervision manner. 
	On real-world RS cases, our SelfDRSC can produce high framerate GS videos with finer correction textures and better temporary consistency. 
	The source code and trained models are made publicly available at {\url{https://github.com/shangwei5/SelfDRSC}}. 
 We also provide an implementation in HUAWEI Mindspore at \url{https://github.com/Hunter-Will/SelfDRSC-mindspore}.
\end{abstract}

\section{Introduction}
Recent years have witnessed an increasing demand for imaging sensors, due to the widespread applications of digital cameras and smartphones. 
Although the Charge-Coupled Device (CCD) has been the dominant technology for imaging sensors, it is recently popular that modern consumer cameras choose the Complementary Metal-Oxide Semiconductor (CMOS) as an alternative due to its many merits, \eg, easy integration with image processing pipeline and communication circuits, and low power consumption~\cite{litwiller2001ccd}.
In CMOS sensors, rolling shutter (RS) scanning mechanism is generally deployed to capture images, \ie, each row of CMOS array is exposed in the sequential time, which is different from CCD with global shutter (GS) scanning at one instant.  
Therefore, RS images suffer from distortions when capturing dynamic scenes, which not only affect human visual perception but also yield performance degradation or even failure in computer vision tasks~\cite{albl2015r6p,lao2020rolling}.

To correct RS distortions, pioneering works usually reconstruct GS images from a single RS image \cite{meingast2005geometric,zhuang2019learning} or multiple consecutive RS images \cite{liu2020deep,Fan_2021_ICCV}, where the latter ones usually have better performance. 
But consecutive RS images setting is ambiguous \cite{zhong2022bringing}, \eg, two RS cameras, moving horizontally at the same speed but with different readout time, can produce the same RS images. 
Most recently, a new RS acquisition setting, \ie, dual RS images with reversed scanning directions (see Fig. \ref{fig:dual RS}), is proposed in \cite{albl2020two,zhong2022bringing} to address this ambiguity. 
In \cite{albl2020two}, one GS image is reconstructed from dual RS images with reversed distortions, while in \cite{zhong2022bringing}, Zhong \etal devote efforts to reconstruct high framerate GS videos.  
Nevertheless, both of them adopt a fully supervised learning manner, \ie, ground-truth GS supervision is required to learn RS correction networks. 
Especially in \cite{zhong2022bringing}, high framerate GS videos should be collected to serve as ground-truth. 
In the supervised learning manner, it is not easy to collect real-world training samples, while synthetic training samples would yield poor generalization ability when handling real-world RS cases.

\begin{figure*}[!t]\footnotesize
	\centering
	\setlength{\abovecaptionskip}{3pt} 
	\setlength{\belowcaptionskip}{0pt}
	\begin{tabular}{cccccc}
		\includegraphics[width=0.95\linewidth]{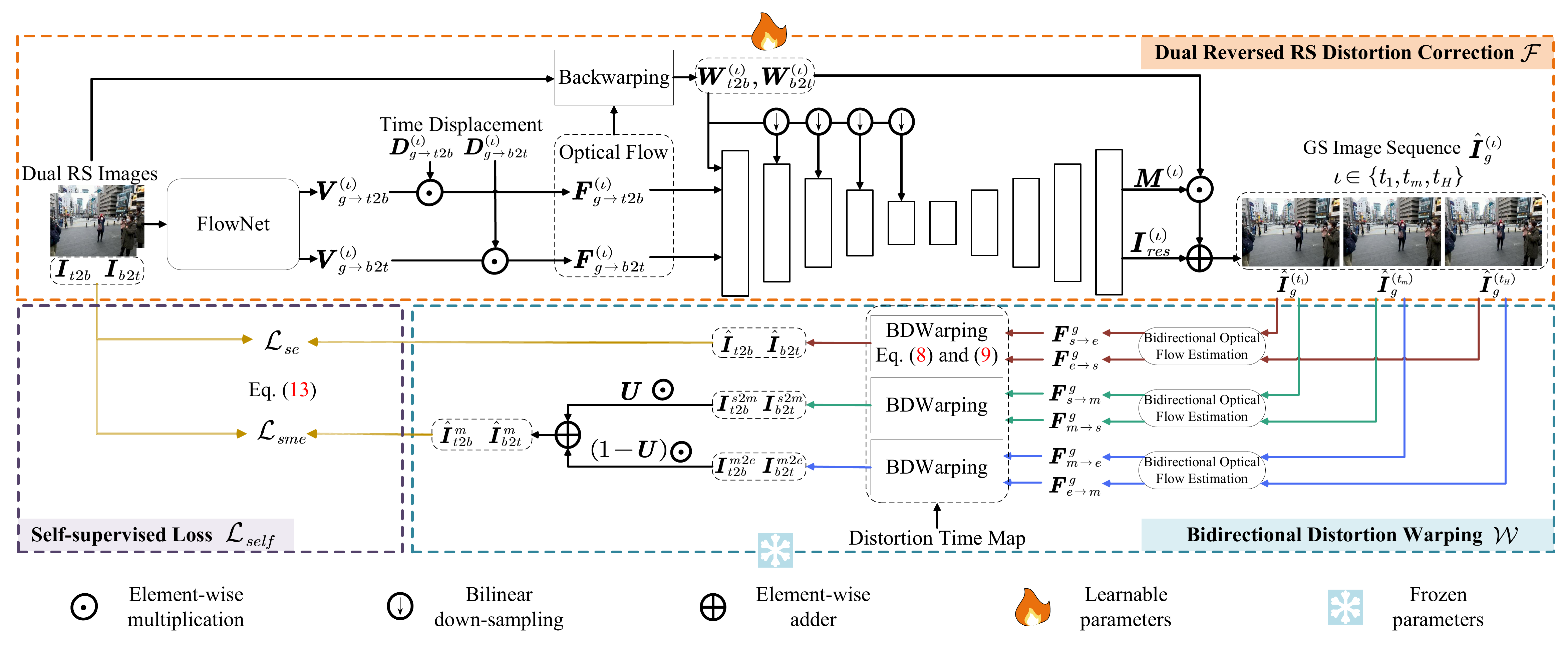}\\
	\end{tabular}
	\caption{Training framework of our SelfDRSC, which consists of three modules, \ie, DRSC network $\mathcal{F}$ for generating GS images $\{\hat{\bm{I}}_g^{(t_1)}, \hat{\bm{I}}_g^{(t_m)}, \hat{\bm{I}}_g^{(t_H)}\}$ from input dual RS images $\bm I_{t2b}$ and $\bm I_{b2t}$, a bidirectional distortion warping module $\mathcal{W}$ for reconstructing dual reversed RS images, and self-supervised loss $\mathcal{L}_{self}$ for enforcing the cycle consistency between input and reconstructed RS images. 
		Moreover, a self-distillation loss $\mathcal{L}_{sd}$, referring to Fig. \ref{fig:self-dis}, is introduced into self-supervised loss for mitigating boundary artifacts in generated GS images.
		In inference phase, the learned DRSC model $\mathcal{F}$ is able to generate high framerate GS videos by giving multiple intermediate time $t_m$.
	}
	\label{fig:framework}
\end{figure*}

In this paper, we aim ambitiously for the more challenging and practical task, \ie, self-supervised learning to invert dual reversed RS images to a high framerate GS video, dubbed SelfDRSC, which to the best of our knowledge is studied for the first time. 
%
The primary design philosophy of our SelfDRSC is that latent GS images predicted by the DRSC network can be used to reconstruct dual RS images with reversed distortions, and then the DRSC network can be learned by enforcing the cycle consistency between input and reconstructed RS images.
As shown in Fig. \ref{fig:framework}, a novel bidirectional warping (BDWarping) module is proposed, by which dual reversed RS images can be reconstructed, and then a self-supervised loss can be deployed to train the DRSC network. 
During training, an intermediate GS image at arbitrary RS scanning time is predicted, and in our BDWarping module, it can also be used to reconstruct another set of dual RS images, which serve as the extra self-supervision in SelfDRSC. 
In this way, the predicted GS images at intermediate scanning time can also be supervised, making the learned DRSC network be able to generate high framerate GS videos. 
Moreover, the DRSC network trained by individual self-supervised loss would yield undesirable boundary artifacts as shown in Fig. \ref{fig:boundary}, and we introduce a self-distillation strategy into self-supervised loss to alleviate this issue, as shown in Fig. \ref{fig:self-dis}.  
Extensive experiments on synthetic and real-world RS images have been conducted to evaluate our SelfDRSC.  
Although any ground-truth GS supervision is not exploited in our SelfDRSC, it still achieves comparable quantitative metrics on synthetic dataset, in comparison to state-of-the-art supervised RS correction methods. 
On real-world RS cases, our SelfDRSC can produce high framerate GS videos with finer textures and better temporary consistency. 
\section{Related Work}
In this section, we provide a brief review on rolling shutter correction methods, and self-supervised methods in low-level vision.  

\subsection{Rolling Shutter Distortion Correction}
With the rising demand for RS cameras, RS distortion corrections have received widespread attention. 
Existing works on RS correction generally fall into two categories: single-image-based~\cite{meingast2005geometric,zhuang2019learning} and multi-frame-based~\cite{liu2020deep,Fan_2021_ICCV,zhong2022bringing} methods. 
It is an ill-posed problem to correct RS distortion from a single image, and its performance is usually inferior. 
%
%
%
For multi-frame-based methods, it can be further divided into generating one specific image and generating a video sequence.
For the former, Liu \etal~\cite{liu2020deep} proposed an end-to-end model, which warped features of RS images to a GS image by a special forward warping block.
For the latter, Fan \etal~\cite{Fan_2021_ICCV} designed a deep learning framework, which utilized the underlying spatio-temperal geometric relationships for generating a latent GS image sequence. Then Fan \etal~\cite{Fan_2022_CVPR} further proposed a context-aware model for solving complex occlusions and object-specific motion artifacts.
Recently, Zhong \etal~\cite{zhong2022bringing} proposed an end-to-end method IFED, and it can extract an undistorted GS sequence grounded on the symmetric and complementary nature of dual RS images with reversed distortion~\cite{albl2020two}.
%
However, these methods all rely on supervised learning, which would yield poor generalization on real-world data.
%
In this work, we aim to develop a self-supervised learning framework for inverting dual reversed RS images to high framerate GS videos with visually pleasing results. 

\subsection{Cycle Consistency-based Self-supervised Learning in Low-level Vision}
%
The concept of cycle consistency has been utilized in several low-level vision tasks for self-supervised learning.
Zhu~\etal~\cite{zhu2017unpaired} proposed a cycle consistency loss for unpaired image-to-image translation.
%
%
Chen \etal~\cite{chen2018reblur2deblur} enforced the results by fine-tuning existing methods in a self-supervised fashion, where they estimated the per-pixel blur kernel based on optical flows between restored frames, for reconstructing blurry inputs.
Ren \etal~\cite{ren2020neural} utilized two generative networks for respectively modeling the deep priors of clean image and blur kernel to reconstruct blurred image for enforcing cycle consistency.
Liu \etal~\cite{liu2020self} claimed that motion cues obtained from consecutive images yield sufficient information for deblurring task and they re-rendered the blurred images with predicting optical flows for cycle consistency-based self-supervised learning.
%
%
Bai \etal~\cite{bai2022self} presented a self-supervised video super-resolution method, which can generate auxiliary paired data from the original low resolution input videos to constrain the network training. 
%
To sum up, in these methods, degraded images can be reconstructed based on the imaging mechanism, and then self-supervised loss can be employed to enforce the cycle consistency between reconstructed and original degraded images. 
In this work, we incorporate cycle consistency-based self-supervised loss with self-distillation for tackling DRSC problem. 



\section{Proposed Method}
In this section, we first present the problem formulation of self-supervised learning for Dual reversed RS Correction (SelfDRSC). 
Then, the DRSC network architecture is briefly introduced, and more attention is paid on our proposed self-supervised learning framework including bidirectional distortion warping module and self-supervised loss with self-distillation. 

\begin{figure}[!t]\footnotesize
	\centering
	\setlength{\abovecaptionskip}{3pt} 
	\setlength{\belowcaptionskip}{0pt}
	\begin{tabular}{cccccc}
		\includegraphics[width=0.95\linewidth]{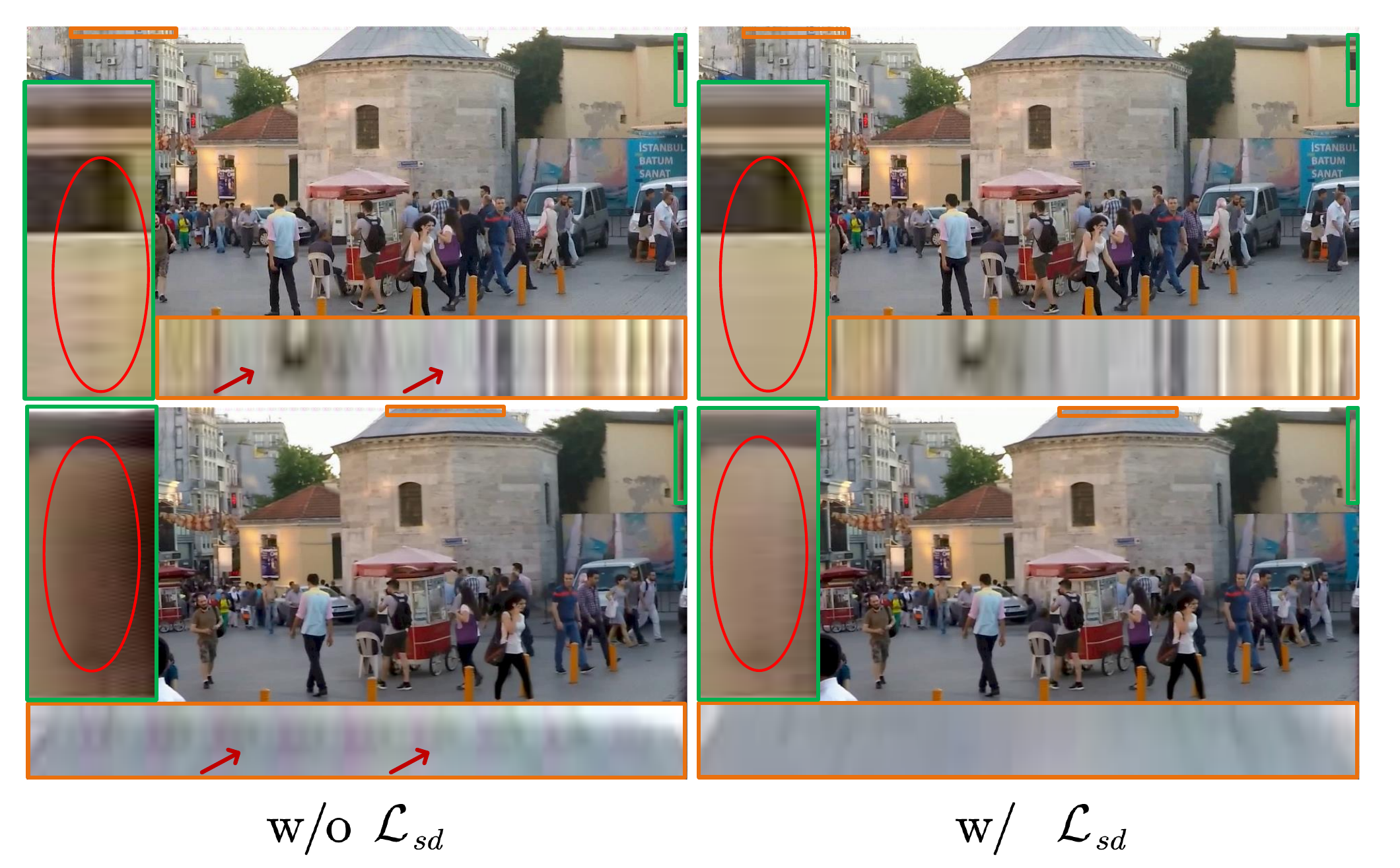}\\
	\end{tabular}
	\caption{Self-distillation $\mathcal{L}_{sd}$ for mitigating boundary artifacts.
	}
	\label{fig:boundary}
\end{figure}
\subsection{Formulation of SelfDRSC}
Recently, dual reversed RS imaging setting~\cite{albl2020two,zhong2022bringing} has been proposed to address the ambiguity issue in consecutive RS images, where two RS images are captured simultaneously by different scanning patterns, \ie, top-to-bottom ($\bm I_{t2b}$) and bottom-to-top ($\bm I_{b2t}$), as shown in Fig. \ref{fig:dual RS}. 
We first give a formal imaging formation of dual RS images with $H$ rows. 
Without loss of generality, we define the acquisition time $t$ as the midpoint of the whole exposure period, \ie, each RS image is captured from $t_1$ to $t_H$, having $H-1$ readout instants $\tau$, where $t_1=t-\tau (H-1)/2$ and $t_H=t+\tau (H-1)/2$.
%
Dual reversed RS images captured at time $t$ can be defined as 
\begin{equation}\label{eq:dual RS}
	\begin{aligned}
		\bm{I}^{(t)}_{t2b}[i] &= \bm{I}_g^{(t+\tau(i-(H+1)/2))}[i], \\
		\bm{I}^{(t)}_{b2t}[i] &= \bm{I}_g^{(t+\tau(i-(H+1)/2))}[H-i+1], 
	\end{aligned}    
\end{equation}
where $\bm{I}_g$ is the latent GS image. 
When scanning $i$-th rows for $\bm{I}^{(t)}_{t2b}$ and $\bm{I}^{(t)}_{b2t}$, the image contents are captured from $\bm{I}_g$ at the same instant time $t_i$ but with reversed scanning directions. 
In the following, the superscripts in $\bm{I}_{t2b}^{(t)}$ and $\bm{I}_{b2t}^{(t)}$ are omitted, since $t=(t_1+t_H)/2$.

Under the dual reversed RS imaging setting, RS distortions can be well distinguished, \ie, dual reversed RS images $\bm{I}^{}_{t2b}$ and $\bm{I}^{}_{b2t}$ provide cues for reconstructing GS images between $t_1$ and $t_H$, even for the ambiguous case in consecutive video setting. 
In the most recent state-of-the-art method IFED \cite{zhong2022bringing}, fully supervised learning is employed for learning DRSC network, where high framerate GS video frames should be collected as ground-truth supervision.  
Albeit obtaining promising performance, high framerate GS frames are not trivial to collect. 
The common solution is to synthesize training datasets where video frame interpolation (VFI) is usually adopted to increase video framerate \cite{zhong2022bringing}, thus restricting their performance on real-world cases.

In this work, we propose a novel SelfDRSC method for rolling shutter correction with dual reversed distortion, where only RS images are required for training DRSC network, without requiring ground-truth high framerate GS images. 
Formally, the optimization of SelfDRSC is defined as
\begin{equation}
	\underset{\bm\Theta}{ \min } \ \mathcal{L}\left(\left\{\bm I_{t 2 b}, \bm I_{b 2 t}^{}\right\}, \mathcal{W}\left(\mathcal{F}\left(\bm I_{t 2 b}^{}, \bm I_{b 2 t}^{}; \bm \Theta \right)\right)\right),
\end{equation}
which contains three key components: DRSC network $\mathcal{F}$ with parameters $\bm{\Theta}$, bidirectional distortion warping (BDWaring) module $\mathcal{W}$, and self-supervised learning objective $\mathcal{L}$. 
By taking dual RS images $\bm{I}_{t2b}^{}$ and $\bm{I}_{b2t}^{}$ as input, DRSC network $\mathcal{F}$ generates high framerate GS frames. 
To learn parameters $\bm{\Theta}$ of $\mathcal{F}$, BDWaring module $\mathcal{W}$ reconstructs dual RS images from generated GS frames, and self-supervised learning loss $\mathcal{L}_{self}$ is imposed to enforce the cycle consistency between input and reconstructed RS images. 
Moreover, self-distillation $\mathcal{L}_{sd}$ is introduced for mitigating boundary artifacts in generated GS images. 


\subsection{Network Architecture of DRSC $\mathcal{F}$ }
%

%
The architecture of $\mathcal{F}$ is similar with that in IFED \cite{zhong2022bringing}, which consists of a RS correction module and a GS reconstruction module. 
In IFED \cite{zhong2022bringing}, multiple GS frames are directly adopted as supervision for training DRSC network, and the output of IFED has fixed framerate. 
In contrast, our SelfDRSC does not require high framerate GS images as ground-truth supervision. 
Therefore, during training phase, our DRSC network $\mathcal{F}$ generates three images $\{\hat{\bm{I}}_g^{(t_1)}, \hat{\bm{I}}_g^{(t_m)}, \hat{\bm{I}}_g^{(t_H)}\}$ where $t_m$ is an intermediate scanning time between start time $t_1$ and end time $t_H$. 
During inference phase, it allows DRSC network to generate videos with arbitrary framerate by giving different intermediate scanning time $t_m$. 
In the following, we take an example to show how intermediate GS image $\hat{\bm{I}}_g^{(t_m)}$ is predicted from dual reversed RS images $\bm{I}_{b2t}$ and $\bm{I}_{t2b}$. 
The start and end GS images $\hat{\bm{I}}_g^{(t_1)}$ and $\hat{\bm{I}}_g^{(t_H)}$ can be obtained in the same way. 

\textit{\textbf{RS Correction Module.}}
For inverting RS images to GS images, a natural strategy is to warp RS images based on the optical flow $\bm F_{g \rightarrow t2b}^{(t_m)}$ and $\bm F_{g \rightarrow b2t}^{(t_m)}$ between input RS images and latent GS images. 
But direct estimation is a challenging problem due to the time displacement and relative motion~\cite{zhong2022bringing}. 
Following IFED, we adopt a simple FlowNet \cite{zhong2022bringing} to estimate relative motion map $\bm V_{g\rightarrow t2b}^{(t_m)}$ and $\bm V_{g\rightarrow b2t}^{(t_m)}$ between latent GS images and the input RS images. 
As for the time displacement $\bm D^{(t_m)}$ between input RS images and latent GS images, they can be obtained based on the scanning mechanism of RS cameras. 
%
%
Formally, the values at $i$-th row in time displacement are given by
\begin{equation}
	\begin{aligned}
		\bm D^{\left(t_m\right)}_{g\rightarrow t2b}[i]&=\frac{i-m}{H-1}, \ &i, m \in [1, \!\cdots,\!H], \\
		\bm D^{\left(t_m\right)}_{g\rightarrow b2t}[i]&=\frac{(H-i)-(m-1)}{H-1}, \ &i, m \in [1,\!\cdots,\!H]. 
	\end{aligned}
\end{equation}
Then the optical flow between input RS images and latent GS images can be obtained by multiplying corresponding time displacement and relative motion map in the entry-by-entry manner, \ie, $\bm{F}_{g\rightarrow t2b}^{(t_m)} = \bm{D}_{g\rightarrow t2b}^{(t_m)} \odot \bm{V}_{g\rightarrow t2b}^{(t_m)}$ and $\bm{F}_{g\rightarrow b2t}^{(t_m)} = \bm{D}_{g\rightarrow b2t}^{(t_m)}\odot \bm{V}_{g\rightarrow b2t}^{(t_m)}$.
Finally, the corrected images $\bm W^{(t_m)}_{t2b}=\mathcal{B}(\bm{I}_{t2b};\bm{F}_{g\rightarrow t2b}^{(t_m)} )$ 
and $\bm W^{(t_m )}_{b2t}=\mathcal{B}(\bm{I}_{b2t};\bm{F}_{g\rightarrow b2t}^{(t_m)})$ can be obtained from $\bm{I}_{t2b}$ and $\bm{I}_{b2t}$ by the backwarping operation $\mathcal{B}$ \cite{zhong2022bringing}. 

\textit{\textbf{GS Reconstruction Module.}}
The remaining issue is how to fuse warped images $\bm W^{(t_m)}_{t2b}$ and $\bm W^{(t_m )}_{b2t}$ for reconstructing latent GS image $\bm{I}_g^{(t_m)}$. 
In our work, encoder-decoder is adopted, where dual corrected images ($\bm{W}^{(t_m)}_{t2b}$ and $\bm{W}^{(t_m)}_{b2t}$) and optical flows ($\bm{F}^{(t_m)}_{g\rightarrow t2b}$ and $\bm{F}^{(t_m)}_{g\rightarrow b2t}$) are taken as input, and a fusing mask $\bm{M}^{(t_m)}$ and residual image $\bm{I}_{res}^{(t_m)}$ are generated as output. 
In particular, the reconstruction is performed in a multi-scale framework, where 5 levels are adopted. 
And finally, the GS image can be obtained by 
\begin{equation}\label{eq:drsc}
	\hat{\bm I}_{g}^{(t_m)}\!\!=\!\bm I_{res}^{(t_m)}\!+\!\bm M^{(t_m)} \!\odot\! \bm W_{t2b}^{(t_m)}\!+\!(\bm 1\!-\!\bm M^{(t_m)}\!) \odot\! \bm W_{b2t}^{(t_m)}.
\end{equation} 
More details of $\mathcal{F}$ can be found in the supplementary file. 
%

%

%
%



\subsection{Self-supervised Learning for DRSC}
%
To learn the parameters $\bm \Theta$ of DRSC network $\mathcal{F}$, we need to introduce supervision on $\{\hat{\bm{I}}_g^{(t_1)}, \hat{\bm{I}}_g^{(t_m)}, \hat{\bm{I}}_g^{(t_H)}\}$. 
Instead of collecting ground-truth GS images, we suggest that supervision can be exploited from the input RS images themselves. 
Generally, we introduce a bidirectional distortion warping module $\mathcal{W}$ to reconstruct dual reversed RS images, and self-supervised loss $\mathcal{L}$ can be adopted to learn the parameters  $\bm \Theta$ without ground-truth GS images.


\vspace{-0.5em}
\subsubsection{Reconstruction of Dual Reversed RS Images}
\vspace{-0.5em}
Based on RS imaging mechanism, generated start and end GS images $\hat{\bm{I}}_g^{(t_1)}$ and $\hat{\bm{I}}_g^{(t_H)}$ can be accordingly exploited to reconstruct dual reversed RS images $\hat{\bm{I}}_{t2b}$ and $\hat{\bm{I}}_{b2t}$. 
Besides start and end GS images, we also provide a way to reconstruct RS images $\hat{\bm{I}}^m_{t2b}$ and $\hat{\bm{I}}^m_{b2t}$ from intermediate GS images $\hat{\bm{I}}_g^{(t_m)}$. 
In the following, we take top-to-bottom scanning pattern as an example to show the reconstruction of $\hat{\bm{I}}_{t2b}$.

\begin{figure}[!t]\footnotesize
	\centering
	\setlength{\abovecaptionskip}{3pt} 
	\setlength{\belowcaptionskip}{0pt}
	\begin{tabular}{cccccc}
		\includegraphics[width=0.9\linewidth]{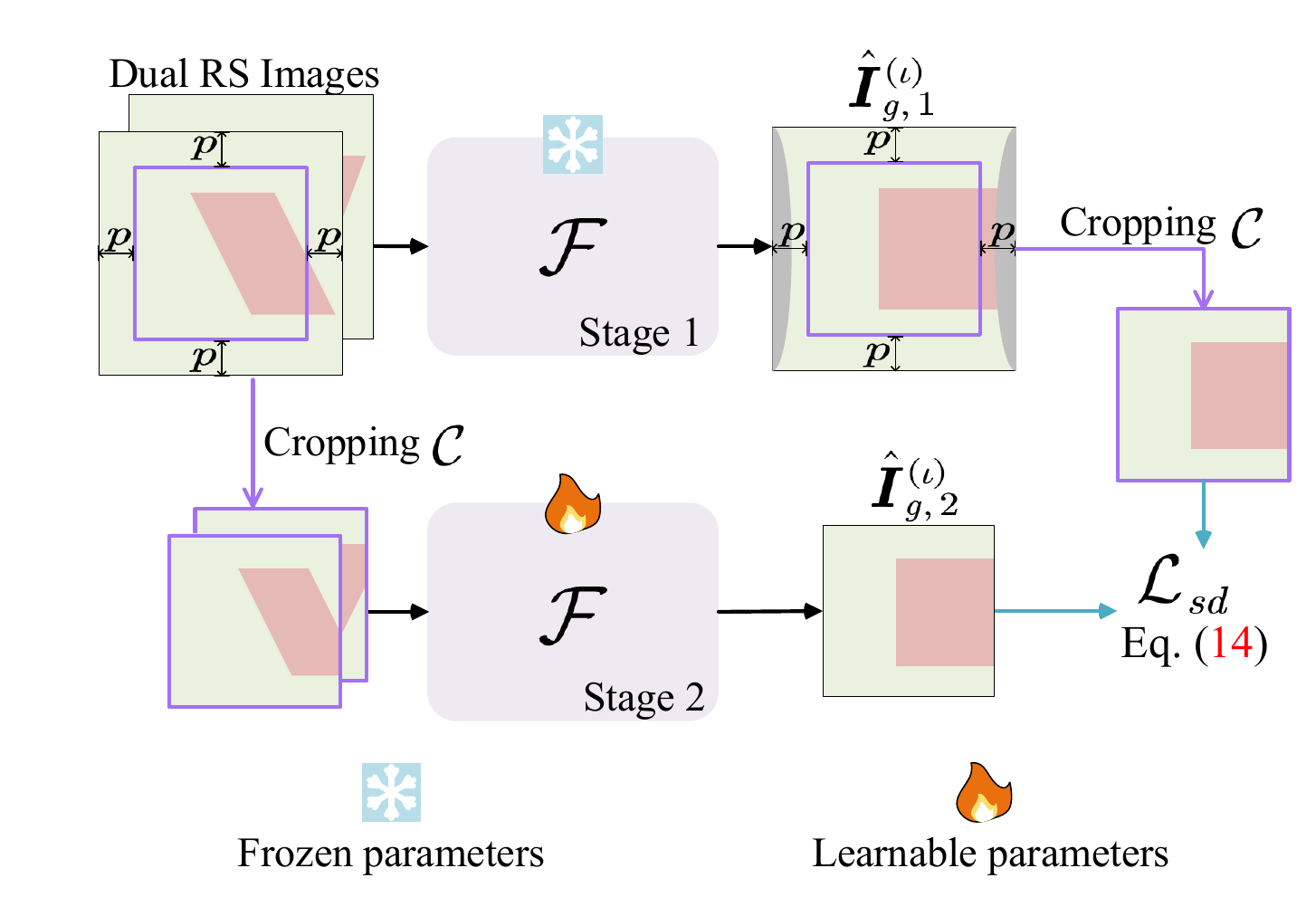}\\
	\end{tabular}
	\caption{
		Self-distillation loss for mitigating boundary artifacts. 
		The DRSC network $\mathcal{F}$ learned by individual $\mathcal{L}_{self}$ in Stage 1 generates GS images $\hat{\bm{I}}_{g,1}^{(\iota)}$ having high-quality center regions but suffering from boundary artifacts. 
		For training $\mathcal{F}$ in Stage 2, the GS images $\hat{\bm{I}}_{g,1}^{(\iota)}$ are cropped by $\mathcal{C}$ with boundary cropping size $p$ to serve as pseudo GS supervision for finetuning $\mathcal{F}$, whose boundary artifacts can be well mitigated. 
	}
	\label{fig:self-dis}
\end{figure}

\textit{\textbf{Reconstructing RS Images from Start \& End GS Frames $\hat{\bm{I}}_g^{(t_1)}$ and $\hat{\bm{I}}_g^{(t_H)}$.}}
The purpose of our bidirectional distortion warping $\mathcal{W}$ is to reconstruct $\hat{\bm{I}}_{t2b}$ from $\hat{\bm I}_{g}^{(t_1)}$ and $\hat{\bm I}_{g}^{(t_H)}$, where the key issue is to obtain bidirectional distortion optical flows $\bm F_{t2b\rightarrow g}^{s2e}$ and $\bm F_{t2b\rightarrow g}^{e2s}$. 
Then $\hat{\bm{I}}_{t2b}$ can be reconstructed using a backwarping operation. 
%
%
%
First, given the corrected results $\hat{\bm I}_{g}^{(t_1)}$ and $\hat{\bm I}_{g}^{(t_H)}$, it is easy to compute bidirectional optical flows ${\bm F}^g_{s \rightarrow e}$ and $\bm F^g_{e \rightarrow s}$ between GS frames, by using a pre-trained optical flow estimation network (\eg, PWC-Net~\cite{sun2018pwc} and GMFlow~\cite{xu2022gmflow}). 
%
%
According to arbitrary time flow interpolation method in VFI \cite{huang2022rife}, we can estimate optical flows between $t_1$ (or $t_H$) and arbitrary intermediate time, \eg, linear approximation~\cite{jiang2018super}, flow reversal~\cite{xu2019quadratic} and CFR~\cite{sim2021xvfi}. 
%
We need to introduce distortion time map in RS imaging process to obtain optical flows $\bm F_{t2b\rightarrow g}^{s2e}$ and $\bm F_{t2b\rightarrow g}^{e2s}$ between GS images and reconstructed RS image $\hat{\bm{I}}_{t2b}$. 

%
Hence, we design distortion time map for representing the interpolation time in each row.
Distortion time map $\bm{T}_{s\rightarrow e}^{t2b}$ for estimating distortion optical flow from start time $t_1$ to end time $t_H$ can be formulated as
\begin{equation}\label{eq:time map}
	\bm{T}_{s \rightarrow e}^{t2b}[i]=\frac{(i-1)\cdot\tau}{(H-1)\cdot\tau}=\frac{i-1}{H-1}, i \in [1, \cdots,H].
\end{equation}
And we can also get $\bm T_{e \rightarrow s}^{t2b}=\bm 1 - \bm T_{s \rightarrow e}^{t2b}$. 
According to CFR~\cite{sim2021xvfi}, we need to first obtain anchor flows $\bm F_{1}^{s2e}$ and $\bm F_{1}^{e2s}$, and complementary flows $\bm F_{2}^{e2s}$ and $\bm F_{2}^{s2e}$ for complementally filling the holes occurred in the reversed flows.
Different from CFR, the time instance at each row is different.
Anchor flows can be calculated as 
\begin{equation}
	\bm F_{1}^{s2e} = \bm{T}_{s\rightarrow e}^{t2b} \odot \bm F_{s\rightarrow e}^{g},\text{ and }
	\bm F_{1}^{e2s} = \bm{T}_{e\rightarrow s}^{t2b} \odot \bm F_{e\rightarrow s}^{g}.
\end{equation}
%
And the complementary flows are normalized as 
\begin{equation}
	\bm F_{2}^{e2s}=\bm{T}_{s\rightarrow e}^{t2b} \odot  \bm F_{e\rightarrow s}^{g},\text{ and }
	\bm F_{2}^{s2e}=\bm{T}_{e\rightarrow s}^{t2b} \odot  \bm F_{s\rightarrow e}^{g}.
\end{equation}
Then we can obtain distortion optical flow as follows
\begin{equation}\label{eq:cfr}
	\small
	\!\!\!\!\! \!
	\begin{aligned}
		\bm F_{t2b\rightarrow g}^{s2e}(\mathbf
		x)\!&=\!\frac{\!\bm T_{s\rightarrow e}^{t2b}(\mathbf
			x)\!\cdot\!\!\sum_{\mathbb{N}_2}\!\!w_2\bm F_{2}^{e2s}\!(\mathbf y_2\!)
			\!-\!\bm T_{e\rightarrow s}^{t2b}(\mathbf
			x)\!\cdot\!\!\sum_{\mathbb{N}_1}\!\!w_1 \bm F_{1}^{s2e}\!(\mathbf
			y_1\!)}
		{\bm T_{e\rightarrow s}^{t2b}(\mathbf
			x)\cdot \sum_{\mathbb{N}_1} \! w_1+\bm T_{s\rightarrow e}^{t2b}(\mathbf
			x)\cdot \sum_{\mathbb{N}_2}\! w_2},\\
		\bm F_{t2b\rightarrow g}^{e2s}(\mathbf
		x)\!&=\!\frac{\bm T_{e\rightarrow s}^{t2b}(\mathbf
			x)\!\cdot\!\!\sum_{\mathbb{N}_1}\!\! w_1\!\bm F_{2}^{s2e}(\mathbf
			y_1\!)\!-\!\bm T_{s\rightarrow e}^{t2b}(\mathbf
			x)\!\cdot\!\sum_{\mathbb{N}_2}\! \! w_2\bm F_{1}^{e2s}\!(\mathbf y_2\!)}{\bm T_{e\rightarrow s}^{t2b}(\mathbf
			x)\cdot \sum_{\mathbb{N}_1} \! w_1+\bm T_{s\rightarrow e}^{t2b}(\mathbf
			x)\cdot \sum_{\mathbb{N}_2} \! w_2},
	\end{aligned}
\end{equation}
where $\mathbf x$ denotes a pixel coordinate, and $\mathbf{y}_1, \mathbf{ y}_2$ are neighbors of $\mathbf x$. 
The neighbors are defined as $\mathbf{y}_1 \in \mathbb{N}_1$ with   $\mathbb{N}_1 = \{\mathbf{y}| \operatorname{round}\left(\mathbf{y}+\bm F_{1}^{s2e}(\mathbf{y})\right)=\mathbf{x}\}$,
and 
$\mathbf{y}_2 \in \mathbb{N}_2$ with   $\mathbb{N}_2 = \{\mathbf{y}| \operatorname{round}\left(\mathbf{y}+\bm F_{1}^{e2s}(\mathbf{y})\right)=\mathbf{x}\}$.
The $\operatorname{round}(\cdot)$ is numerical rounding operator.
%
The Gaussian weights $w_1 = \mathcal{G}(|\mathbf x - (\mathbf y_1 + \bm F_{1}^{s2e}(\mathbf y_1))|)$ and $w_2 = \mathcal{G}( |\mathbf x - (\mathbf y_2 + \bm F_{1}^{e2s}(\mathbf y_2))|)$ are depending on the distance between pixel coordinates.
%
Finally, the RS image $\hat{\bm{I}}_{t2b}$ is reconstructed as 
\begin{equation}\label{eq:backwarp}
	\hat{\bm{I}}_{t2b}\!=\!\bm{T}_{e\rightarrow s}^{t2b}\odot \mathcal{B}(\hat{\bm I}_{g}^{(t_1)}; \!\bm F_{t2b\rightarrow g}^{s2e})
	\!+\!
	\bm{T}_{s\rightarrow e}^{t2b}\odot \mathcal{B}(\hat{\bm I}_{g}^{(t_H)}; \! \bm F_{t2b\rightarrow g}^{e2s}),
\end{equation}
where $\mathcal{B}$ is the backwarping operation.

%
%

\textbf{\textit{Reconstructing RS Images from Intermediate GS Frame $\hat{\bm{I}}_g^{(t_m)}$.}}
Aiming to make SelfDRSC be able to generate GS frames at time of arbitrary scanline, we also need to constrain the generated intermediate GS image $\hat{\bm{I}}_g^{(t_m)}$ in the similar way. 
%
The only distinction is that bidirectional distortion warping is divided into two parts, \ie, one is from $t_1$ to $t_m$ and the other one is from $t_m$ to $t_H$.
According to Eq.~\eqref{eq:time map}, we can obtain distortion time map $\bm T_{s\rightarrow m}^{t2b}$ and $\bm T_{m \rightarrow e}^{t2b}$ as follows
\begin{equation}
	\begin{aligned}
		\bm{T}_{s\rightarrow m}^{t2b}[i]&=\left\{\begin{array}{ll}
			\frac{i-1}{m-1}, &\quad \; i \in[1, \cdots,m] ,\\
			1, &\quad \; i \in[m+1, \cdots,H] ,
		\end{array}\right. 
		\\
		\bm{T}_{m \rightarrow e}^{t2b}[i]&=\left\{\begin{array}{ll}
			0, & i \in[1, \cdots,m], \\
			\frac{i-m-1}{H-m-1}, & i \in[m+1, \cdots,H].
		\end{array}\right.
	\end{aligned}
\end{equation}
Then we can obtain $\bm I_{t2b}^{s2m}$ and $\bm I_{t2b}^{m2e}$ similar to Eq.~\eqref{eq:cfr} and Eq.~\eqref{eq:backwarp}.
Finally, we can get the reconstructed RS frame $\hat{\bm I}_{t2b}^{m}$ with time mask $\bm U_{t2b}$
\begin{equation}
	\hat{\bm I}_{t2b}^{m}=\bm U_{t2b}\odot \bm I_{t2b}^{s2m}+(\bm 1-\bm U_{t2b})\odot \bm I_{t2b}^{m2e},
\end{equation}
where $\bm U_{t2b}[i]=\left\{\begin{array}{ll}
	0, & \text{if  } i\textgreater m \\
	1, & \text{else}
\end{array}\right.$. 
%
\vspace{-0.5em}
\subsubsection{Self-supervised and Self-distillation Losses}
\vspace{-0.5em}
We have reconstructed two sets of dual reversed RS images, \ie, $\hat{\bm{I}}_{t2b}$, $\hat{\bm{I}}_{b2t}$ and $\hat{\bm{I}}^m_{t2b}$, $\hat{\bm{I}}^m_{b2t}$, and thus a loss function $\ell$ can be imposed on their corresponding input RS images. 
The self-supervised loss function is defined as
\begin{equation}
	\mathcal{L}_{self}=\mathcal{L}_{se}+\mathcal{L}_{sme},
\end{equation}
with 
\begin{equation}\label{eq:loss}
	\begin{aligned}
		\mathcal{L}_{se}&=\ell(\hat{\bm{I}}_{t2b}, \bm{I}_{t2b})+\ell(\hat{\bm{I}}_{b2t}, \bm{I}_{b2t}^{}),
		\\
		\mathcal{L}_{sme}&=\ell(\hat{\bm{I}}_{t2b}^{m}, \bm{I}_{t2b}^{})+\ell(\hat{\bm{I}}_{b2t}^{m}, \bm{I}_{b2t}^{}),
	\end{aligned}	
\end{equation}
%
where $\ell$ is the combination of Charbonnier loss~\cite{lai2018fast} and perceptual loss~\cite{johnson2016perceptual} in our experiments, and the hyper-parameters for balancing them are set as 1 and 0.1. 
%

However, the DRSC model $\mathcal{F}$ trained only with $\mathcal{L}_{self}$ suffers from boundary artifacts (referring to the left part of Fig.~\ref{fig:boundary}). 
This is because boundary pixels are not reliable when reconstructing RS images, and the individual self-supervised loss $\mathcal{L}_{self}$ actually provides invalid supervision for boundary regions. 
%
Luckily, we find that the center part of corrected GS images is free from artifacts, and we further propose to introduce self-distillation into loss function for mitigating boundary artifacts. 
In particular, the training of SelfDRSC is performed in multiple stages, where $\mathcal{L}_{self}$ is first adopted to train $\mathcal{F}$ in Stage 1, and in the following stages, a self-distillation loss~\cite{grill2020bootstrap} is added to cooperate with $\mathcal{L}_{self}$.  
As shown in Fig. \ref{fig:self-dis}, by taking Stage 2 as an example, $\mathcal{F}$ from Stage 1 has been able to generate GS images having high-quality center regions. 
We assume generated GS images $\{\hat{\bm{I}}_{g,1}^{(t_1)},\hat{\bm{I}}_{g,1}^{(t_m)},\hat{\bm{I}}_{g,1}^{(t_H)}\}$ from Stage 1 have spatial size $H \times H$. 
When training $\mathcal{F}$ at Stage 2, we adopt a cropping operation to extract the center region of input RS images, and $\{\hat{\bm{I}}_{g,1}^{(t_1)},\hat{\bm{I}}_{g,1}^{(t_m)},\hat{\bm{I}}_{g,1}^{(t_H)}\}$ are also cropped in the same way to serve as pseudo GS images
\begin{equation}
	\mathcal{L}_{sd}=\sum_{\iota \in {\{t_1,t_m,t_H\}}}\ell( \hat{\bm{I}}_{g,2}^{(\iota)}, \mathcal{C}(\hat{\bm{I}}_{g,1}^{(\iota)})),
\end{equation}
where $\mathcal{C}$ is boundary cropping operation with size $p$, and the generated GS images $\{\hat{\bm{I}}_{g,2}^{(t_1)},\hat{\bm{I}}_{g,2}^{(t_m)},\hat{\bm{I}}_{g,2}^{(t_H)}\}$ in Stage 2 have spatial size $(H-2p)\times (H-2p)$.
In SelfDRSC, the final loss function $\mathcal{L}$ is defined as 
\begin{equation}
	\mathcal{L}=\left\{\begin{array}{ll}
			\mathcal{L}_{self}, & \text{  when stage $n=1$} \\
			\mathcal{L}_{self}+\mathcal{L}_{sd}, & \text{  when stage } n= 2,...,N
	\end{array}\right.
\end{equation}
and we empirically find that $N=2$, \ie, self-distillation one time, is sufficient to mitigate boundary artifacts. 
\section{Experiments}

In this section, our SelfDRSC is evaluated on synthetic and real-world RS images, and more video results can be found from the link \href{https://1drv.ms/u/s!Aq_1EOtIk78OeP9qJ5_wg8WUU9k?e=dwUgsp}{\textbf{\emph{Video Results}}}. 

\subsection{Datasets}
\vspace{-0.5em}
\subsubsection{Synthetic Dataset}\label{sec:data}
\begin{figure}[h]\footnotesize
	\setlength{\abovecaptionskip}{0.in}
	\setlength{\belowcaptionskip}{0.in}
	\setlength{\tabcolsep}{1.5pt}
	\begin{tabular}{ccccc}
		\includegraphics[width=0.33\linewidth]{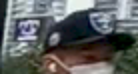}&  
		\includegraphics[width=0.33\linewidth]{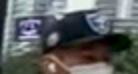}&
		\includegraphics[width=0.33\linewidth]{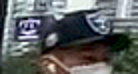}\\
		\multicolumn{1}{c}{\scriptsize GT (PSNR$\uparrow$/SSIM$\uparrow$/LPIPS$\downarrow$)} &
		\multicolumn{1}{c}{\scriptsize IFED (22.23/0.711/0.105)} &
		\multicolumn{1}{c}{\scriptsize Ours (20.01/0.656/0.131)}  \\
	\end{tabular}
	\caption{Interpolated GT frames yield over-smoothed details.  
		The result by IFED has better FR-IQA metrics than Ours, but it has less texture details. 
	}
	\label{fig:motion}
	\vspace{-0.1in}
\end{figure}

\vspace{-0.5em}
The synthetic dataset RS-GOPRO from IFED \cite{zhong2022bringing} is used for quantitatively evaluating the competing methods. 
The training, validation and testing sets are randomly split to have 50, 13 and 13 sequences. 
For training and testing in IFED \cite{zhong2022bringing}, 9 GS images are used as ground-truth. 
We note that ground-truth is not entirely captured by a GOPRO camera, but is partially synthesized by video interpolation methods, possibly yielding over-smoothed details as shown in Fig. \ref{fig:motion}.  
Thus full-reference image assessment (FR-IQA) metrics are not very trustworthy for this task, where the result by IFED has less textures than ours, but is better in FR-IQA metrics. 
\vspace{-0.1in}
%
%

\begin{table*}[!htb]\footnotesize  
	\setlength{\tabcolsep}{5pt}
	\centering
	\setlength{\abovecaptionskip}{0pt} 
	\setlength{\belowcaptionskip}{0pt}
	\begin{tabu}{c|c|c|c|c|cc}
		\hline
		
		\hline	
		& Method &  Inference time (\textit{s})  & Parameters &  NIQE$\downarrow$ / NRQM$\uparrow$ / PI$\downarrow$ &  PSNR$\uparrow$ / SSIM$\uparrow$ / LPIPS$\downarrow$&  \\
		\hline
		\multirow{5}{*}{{Full-supervised}} & DUN+RIFE & 0.638  &   14.62\text{M}  &  3.836 / 6.486 / 3.433  &  23.597 / 0.7653 / 0.1670& \\  
		& RIFE+DUN & 4.078  & 14.62\text{M}  &  3.844 / 6.398 / 3.487  &  20.012 / 0.6520 / 0.1781 & \\  
		&  RSSR & 0.976   & 26.03\text{M}  &  \textbf{3.293} / \underline{6.826} / \underline{2.963}  &  22.729 / 0.7283 / 0.1026 & \\   
		&  CVR & 2.088  & 42.70\text{M}  &  3.667 / 6.420 / 3.348  & 24.816 / 0.7804 / 0.0738 & \\   
		\cline{2-7}
		&  IFED &  0.177& 29.86\text{M}  &  3.657 / 6.405 / 3.358 & \textbf{30.681} / \textbf{0.9121} / \textbf{0.0453} &(*Upper Bound)   \\   
		\tabucline[1pt]{-}
		
		Self-supervised &SelfDRSC (Ours) & 0.182 & 28.75\text{M} &  \underline{3.297} / \textbf{6.933} / \textbf{2.896} &  \underline{28.704} / \underline{0.8886} / \underline{0.0546} &   \\ 
		
		
		\tabucline[1pt]{-}
		Ground-truth &  --- & --- & ---  & 3.506 / 6.639 / 3.159 & --- &  \\
		
		\hline
		
		\hline
	\end{tabu}
	\caption{Quantitative comparison on RS-GOPRO dataset \cite{zhong2022bringing}, where the metrics are computed based on 9 GS images for a testing case.
		*Considering that our SelfDRSC has the similar DRSC network with IFED \cite{zhong2022bringing}, these FR-IQA metrics PSNR, SSIM and LPIPS by IFED can be treated as the upper bound. 
		Also, interpolated GS images by VFI method \cite{huang2022rife} may appear in 9 ground-truth GS images on RS-GOPRO dataset, \ie, ground-truth GS images for computing FR-IQA metrics may not be truly captured by a GOPRO camera, making FR-IQA metrics not so reliable that we suggest further referring to NR-IQA metrics and visual comparison. 
	}\label{tabel:syn}
\end{table*}
\begin{figure*}[!t]\footnotesize
	\centering
	\setlength{\abovecaptionskip}{3pt} 
	\setlength{\belowcaptionskip}{0pt}
	\begin{tabular}{cccccc}
		\includegraphics[width=0.95\linewidth]{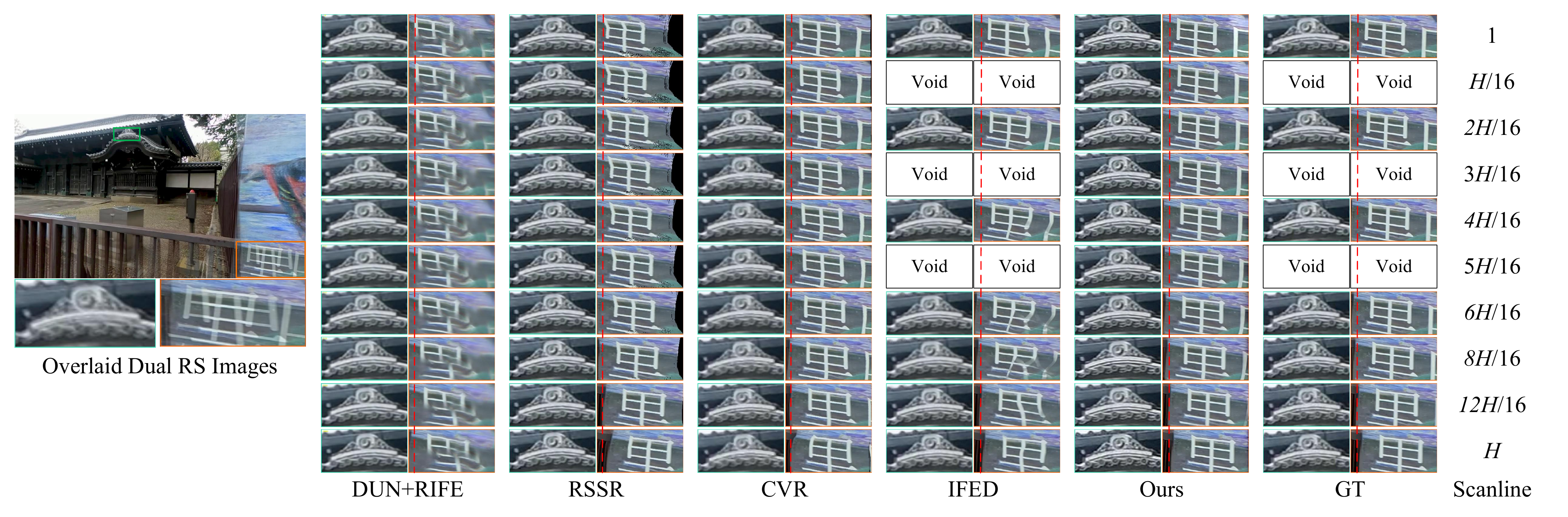}\\
	\end{tabular}
	\caption{Visual results on RS-GOPRO.
		IFED \cite{zhong2022bringing} can only generate a GS video with 9 frames, since 9 ground-truth GS images from RS-GOPRO are used to train DRSC network in a supervised manner. 
		Our SelfDRSC is able to generate GS videos with higher framerate. 
		In this case, 17 GS frames are generated by SelfDRSC. 
		It will be better viewed by zooming in. }
	\label{fig:syn}
\end{figure*}
\vspace{-0.5em}
\subsubsection{Real-world Testing Set}
\vspace{-0.5em}
For testing in real-world, Zhong \etal~\cite{zhong2022bringing} built a dual-RS image acquisition system, which consists of a beam-splitter and two RS cameras with dual reversed scanning patterns. 
%
The readout setting of the dual-RS system can be changed by replacing the type of RS camera (\eg, FL3-U3-13S2C, BFS-U3-63S4C). 
%
Each sample includes two RS distorted images with reversed distortions but without a corresponding ground-truth sequence.
%

\subsection{Implementation Details}
Our SelfDRSC is trained in $N=2$ stages, where self-distillation is introduced in Stage 2. 
The training is done in 205K iterations for Stage 1 and in 115K iterations for Stage 2. 
The optimization is implemented using AdamW~\cite{loshchilov2017decoupled} optimizer ($\beta_1$=0.9, $\beta_2$=0.999) and the initial learning rate is $1 \times 10^{-4}$, which is gradually reduced to $1 \times 10^{-6}$ with the cosine annealing~\cite{loshchilov2016sgdr}.
%
In Stage 1, patch size is $256\times 256$, and batch size is 48. 
In Stage 2, patch size is increased to $320\times 320$ considering that the boundary cropping size is $p = 32$, and batch size is 32. 
%
%
%
During training, the intermediate time $t_m$ is randomly sampled between start time $t_1$ and end time $t_H$ with sampling interval $(t_H-t_1)/8$. 

\subsection{Comparison with State-of-the-art Methods}
To keep consistent quantitative evaluation with IFED \cite{zhong2022bringing}, the output of competing methods should have 9 GS images. 
Thus, our SelfDRSC is compared with methods from two categories. 
(i) The first category contains cascade methods, where RS correction method DUN~\cite{liu2020deep} for generating one GS image and a VFI model RIFE \cite{huang2022rife} is adopted for interpolating 8 GS images.
%
Both of them are re-trained on RS-GOPRO dataset for a fair comparison. 
In Table \ref{tabel:syn}, both cascade orders are considered, \ie, DUN+RIFE and RIFE+DUN. 
%
%
(ii) The second category contains RS correction methods with multiple output images. 
There are only two works by adopting the dual reversed RS correction setting, \ie, IFED \cite{zhong2022bringing} and Albl \etal \cite{albl2020two}. 
Since the source code or experiment results of Albl \etal \cite{albl2020two} are not publicly available, it is not included into comparison. 
Besides, we take RSSR~\cite{Fan_2021_ICCV} and CVR~\cite{Fan_2022_CVPR} into comparison, which are developed for correcting RS distortions from consecutive two RS images with only top-to-bottom scanning. 
For a fair comparison, they are re-trained based on the RS-GOPRO dataset. 
%
%
%
%
As for quantitative metrics, both FR-IQA (\ie, PSNR, SSIM~\cite{wang2004image} and LPIPS~\cite{zhang2018unreasonable}) and NR-IQA (\ie, NIQE~\cite{mittal2012making}, NRQM~\cite{ma2017learning} and PI~\cite{blau20182018})  metrics  are employed to evaluate the competing methods. 

\begin{figure*}[!t]\footnotesize
	\setlength{\abovecaptionskip}{3pt} 
	\setlength{\belowcaptionskip}{0pt}
	\hspace{-1em}
	\begin{tabular}{l}
		\includegraphics[width=0.33\linewidth]{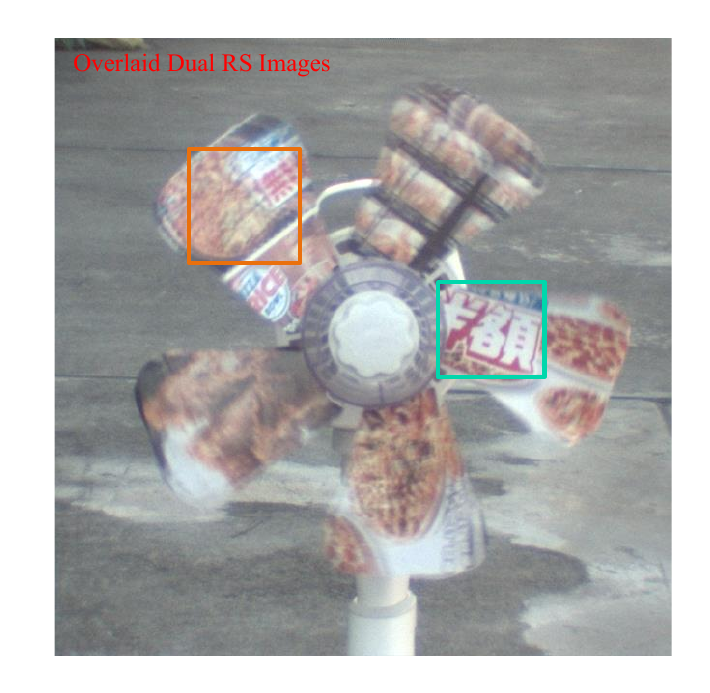}  
		 \animategraphics[width=0.33\linewidth, autoplay, loop]{10}{fig/crop1/}{1}{17}
		 \animategraphics[width=0.33\linewidth, autoplay, loop]{10}{fig/crop2/}{1}{17}		
		 \\
	\end{tabular}
	\caption{Video results on real-world RS data. The animated videos can be viewed in Adobe PDF reader.}
	\label{fig:real}
\end{figure*}
\vspace{-0.5em}
\subsubsection{Results on Synthetic Dataset}
\vspace{-0.5em}
Table \ref{tabel:syn} reports quantitative comparison, where the IQA metrics are computed based on 9 GS images for each testing case. 
Our SelfDRSC, without any ground-truth high framerate GS images during training, achieves better FR-IQA metrics than the supervised methods except IFED. 
Considering that our SelfDRSC has similar DRSC network architecture with IFED \cite{zhong2022bringing}, these FR-IQA metrics PSNR, SSIM and LPIPS by IFED can be treated as the upper bound of self-supervised DRSC methods on RS-GOPRO dataset. 
Due to supervised learning on specific synthetic setting, IFED also has a worse generalization on different camera parameters (\eg, readout time $\tau$ for each row), referring to  \href{https://1drv.ms/u/s!Aq_1EOtIk78OeP9qJ5_wg8WUU9k?e=dwUgsp}{\textbf{\emph{Video Results}}}. 
Besides, as mentioned above, interpolated frames by VFI method may appear in 9 ground-truth GS images for calculating FR-IQA values. 
And thus FR-IQA metrics may not be so reliable to indicate correction performance (please see Fig. \ref{fig:motion}). 
We suggest further referring to NR-IQA metrics and visual results for evaluating the competing methods. 

%
%
Besides FR-IQA metrics, NR-IQA metrics are also employed as reference for quantitative evaluation, and more visual results especially videos are provided for comprehensive justification. 
%
We note that our SelfDRSC is better than ground-truth GS images in terms of NR-IQA values, since they may be interpolated using VFI method rather than original GS frames captured by GOPRO camera. 
As for the visual quality, reconstructed GS images by competing methods are presented in Fig. \ref{fig:syn}. 
IFED \cite{zhong2022bringing} can only generate a GS video with 9 frames. 
Our SelfDRSC is able to generate GS videos with higher framerate. 
In this case, 17 GS frames are generated by SelfDRSC, and are better corrected with finer textures. 
Also, our SelfDRSC ranks as top-2 efficient method in terms of inference time.

\subsubsection{Results on Real-world Data}
\vspace{-0.5em}
In Fig.~\ref{fig:real}, we provide an example on real-world dual reversed RS images. 
It is an animated figure to compare the video results by competing methods. 
%
%
RSSR and CVR cannot generalize well to real-world motion, and IFED can largely correct RS distortions but it is not as good as our SelfDRSC in terms of both textures and temporal consistency.   
We also present NR-IQA metrics on the Zhong~\etal's dataset in Table \ref{tabel:real}.
%
More results on real-world data are available at the link \href{https://1drv.ms/u/s!Aq_1EOtIk78OeP9qJ5_wg8WUU9k?e=dwUgsp}{\textbf{\emph{Video Results}}}.
\begin{table}[!htb]\footnotesize  
	\setlength{\tabcolsep}{5pt}
	\centering
	\setlength{\abovecaptionskip}{0pt} 
	\setlength{\belowcaptionskip}{0pt}
	\begin{tabu}{c|c|c}
		\hline
		
		\hline	
		& Method &  NIQE$\downarrow$ / NRQM$\uparrow$ / PI$\downarrow$   \\
		
		\hline
		\multirow{3}{*}{{Full-supervised}} &  RSSR  &  \underline{4.495} / \underline{6.958} / \underline{3.794}   \\   
		&  CVR &  4.950 / 5.933 / 4.599 \\   
		&  IFED &  4.873 / 5.953 / 4.611   \\   
		\hline
		Self-supervised &SelfDRSC (Ours) &  \textbf{4.391} / \textbf{7.683} / \textbf{3.646}    \\

		\hline
		
		\hline
	\end{tabu}
	\caption{Quantitative comparison on on real-world RS data \cite{zhong2022bringing}.
	}\label{tabel:real}
\end{table}
\begin{table}[!t]\footnotesize  
	\centering
	\setlength{\abovecaptionskip}{0pt} 
	\setlength{\belowcaptionskip}{0pt}
	\begin{tabular}{c|c|c}
		\hline
		
		\hline	
		&  NIQE$\downarrow$ / NRQM$\uparrow$ / PI$\downarrow$  &  PSNR$\uparrow$ / SSIM$\uparrow$ / LPIPS$\downarrow$  \\
		
		\hline
		\multicolumn{3}{l}{\textbf{Optical flow estimation in} $\mathcal{W}$:} \\
		
		\hline
		GMFlow-\emph{kitti} &   3.221 / 6.958 / 2.845  &  25.314 / 0.8080 / 0.0762  \\
		GMFlow-\emph{sintel} &  3.244 / 6.983 / 2.845  &  25.670 / 0.8221 / 0.0763 \\
		RIFE-Flow &  3.570 / 6.735 / 3.144  &  25.539 / 0.8105 / 0.0653  \\
		PWC-Net  &   3.304 / 6.931 / 2.902 & 28.411 / 0.8848 / 0.0574  \\
		
		\hline
		\multicolumn{3}{l}{\textbf{Warping strategy in} $\mathcal{W}$:} \\
		
		\hline
		L.A. \& S. &  3.303 / 6.909 / 2.915  &  26.267 / 0.8466 / 0.0606  \\
		F. R. \& B.  &  3.303 / 6.907 / 2.911 &  27.175 / 0.8665 / 0.0548    \\
		CFR \& B. &  3.304 / 6.931 / 2.902  &  28.411 / 0.8848 / 0.0574  \\
		
		\hline
		\multicolumn{3}{l}{\textbf{Loss function in $\mathcal{L}$}:} \\

		\hline
		$\mathcal{L}_{se}$   & 11.261 / 7.079 / 6.783  & 20.643 / 0.4575 / 0.1888    \\
		$\mathcal{L}_{sme}$  &  3.246 / 6.871 / 2.909 &  23.241 / 0.7514 / 0.0920   \\
		$\mathcal{L}_{self}$  &  3.304 / 6.931 / 2.902 &  28.411 / 0.8848 / 0.0574  \\
		$\mathcal{L}_{self}+\mathcal{L}_{sd}$  &  3.297 / 6.933 / 2.896 &  28.704 / 0.8886 / 0.0546  \\
		
		%
		
		\hline
		
		\hline
	\end{tabular}
	\caption{Ablation results of SelfDRSC on RS-GOPRO dataset.}\label{tabel:abla}
\end{table}
\vspace{-0.5em}
\subsection{Ablation Study}\label{sec:abl}
To demonstrate the effectiveness of different optical flow estimation methods, warping ways, and loss functions.
%
We implement ablation study on these elements. All the IQA metrics are computed based on 9 GS images.
(i) For optical flow estimation method in $\mathcal{W}$, we use pre-trained RIFE-flow~\cite{huang2022rife}, PWC-Net~\cite{sun2018pwc}, GMFlow-\emph{kitti} and GMFlow-\emph{sintel}~\cite{xu2022gmflow}.
We train all these variants with the same warping strategy and without $\mathcal{L}_{sd}$ for fair comparison on RS-GOPRO dataset. From Table~\ref{tabel:abla}, PWC-Net achieves the best performance on self-supervised dual reversed rolling shutter correction task. Hence, we use PWC-Net as our optical flow estimation method in the following.
(ii) For the ways of distortion warping in $\mathcal{W}$, we use three combinations, \ie,  linear approximation~\cite{jiang2018super} \& splatting~\cite{niklaus2020softmax}, flow reversal~\cite{xu2019quadratic} \& backwarping and CFR~\cite{sim2021xvfi} \& backwarping, which are abbreviated as L.A. \& S., F.R. \& B. and CFR \& B., respectively.  
From Table~\ref{tabel:abla}, the third strategy gets the best performance since it combines the advantages of both linear approximation and flow reversal.
(iii) We also train our method with different loss functions. 
We find that $\mathcal{L}_{sd}$ not only achieves about \emph{+0.3dB} PSNR gains but also alleviates boundary artifacts in visual results in Fig.~\ref{fig:boundary}. Moreover, we verify that $\mathcal{L}_{se}$ and $\mathcal{L}_{sme}$ are both very important for self-supervised RS correction.
%
%
For the update strategy of teacher model in self-distillation, we implement different momentum coefficients $c\!\in\!\{0.9,0.99,0.999,1\}$, and we also set different boundary cropping sizes $p\!\in\!\{16,32,64\}$, and different training stage numbers $N$ in the supplementary file. 
%

\section{Conclusion}
In this paper, we proposed a self-supervised learning framework for correcting dual reversed RS distortions, and high framerate GS videos can be generated by our SelfDRSC. 
In SelfDRSC, a novel bidirectional distortion warping module is proposed to obtain reconstructed dual reversed RS images that can be employed as cycle consistency-based supervision.  
The self-supervised learning loss with self-distillation is proposed for training DRSC network, where self-distillation is effective in mitigating boundary artifacts in generated GS images. 
Extensive experiments have been conducted to validate the effectiveness and generalization ability of our SelfDRSC on both synthetic and real-world data. 

\section*{Acknowledgements}
	This work was supported in part by the National Key Research and Development Project (2022YFA1004100), the National Natural Science Foundation of China (62172127 and U22B2035), the Natural Science Foundation of Heilongjiang Province (YQ2022F004), the Hong Kong ITC Innovation and Technology Fund (9440288), and the CAAI-Huawei MindSpore Open Fund.

\appendix
\section{RS Correction Ambiguity in RS Acquisition Setting of Capturing Consecutive Frames}
\begin{figure}[!t]\footnotesize
	\centering
	\setlength{\abovecaptionskip}{3pt} 
	\setlength{\belowcaptionskip}{0pt}
	\begin{tabular}{c}
		\hspace{-2em}
		\includegraphics[width=1\linewidth]{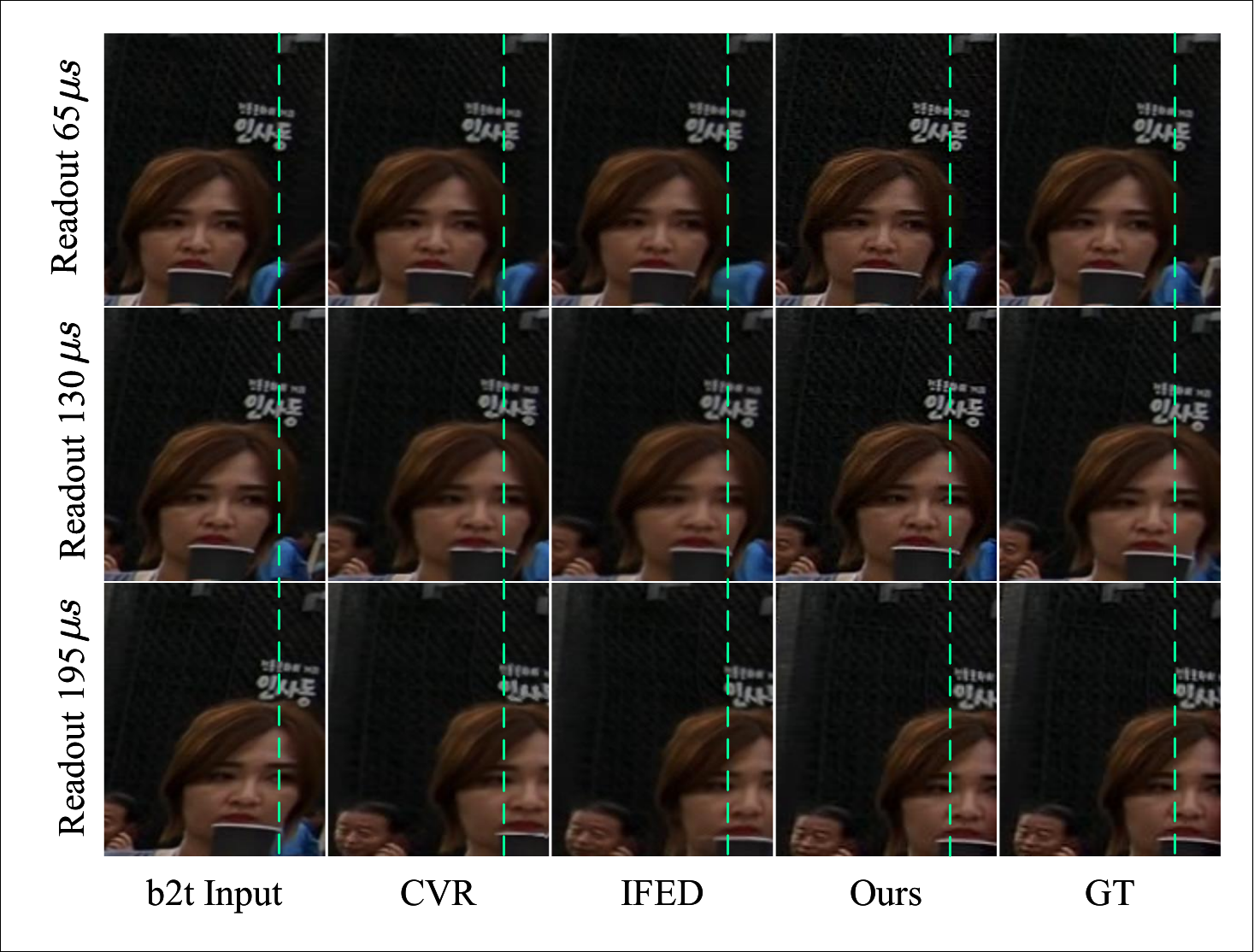}\\
	\end{tabular}
	\vspace{-0.5em}
	\caption{Generalization ability on different readout $\tau$. All methods are trained on fixed readout $\tau=87 \mu s$, while our SelfDRSC can generalize better than others under different readouts.}
	\label{fig:readout}
\end{figure}

\begin{figure*}[!t]\footnotesize
	\centering
	\setlength{\abovecaptionskip}{3pt} 
	\setlength{\belowcaptionskip}{0pt}
	\begin{tabular}{cccccc}
		\includegraphics[width=0.9\linewidth]{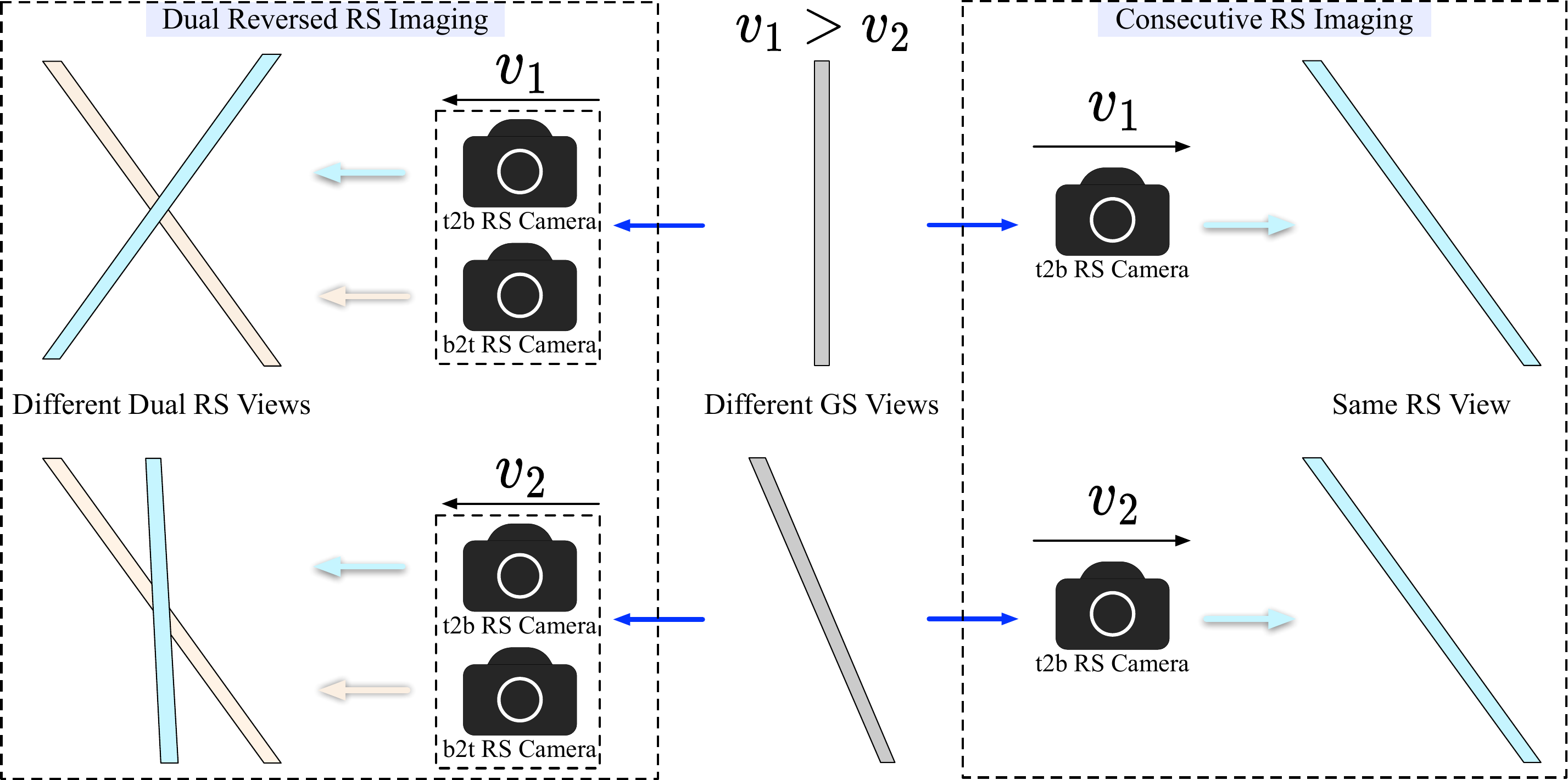}\\
	\end{tabular}
	\caption{
		RS correction ambiguity in RS acquisition setting of capturing consecutive frames. And the acquisition setting of dual RS images with reversed scanning directions can avoid this ambiguity.
	}
	\label{fig:rod}
\end{figure*}

We use a rod as an example in Fig. \ref{fig:rod} to clearly present the ambiguity of in RS acquisition setting of capturing consecutive frames.
Suppose there are two similar rods, one of them is tilted, as shown in GS view. Then, two RS cameras moving horizontally at different speeds $v_1$ and $v_2$ can produce the same RS view, i.e., a fast-moving RS camera with speed $v_1$ for the tilted rod and a slow-moving RS camera with speed $v_2$ for the vertical rod. 
Similarly, two RS cameras moving horizontally at the same speed but with different readout times can also produce the same RS view, i.e., a short readout time RS camera for the tilted rod and a long readout time RS camera for the vertical rod. 
Therefore, the models based on consecutive frames are biased to the training dataset. 
When they face different readout times or speeds from the training data, they become less robust and achieve poor results.
In contrast, the acquisition setting of dual RS images with reversed scanning directions can avoid this ambiguity. However, in state-of-the-art IFED \cite{zhong2022bringing} trained in the full supervision manner, ground-truth high framerate GS images are not truly captured by a GOPRO camera but interpolated frames. 
Therefore, the learned IFED suffers from the loss of textures in the corrected results (see hair and characters in Fig. \ref{fig:readout}), while our method presents good generalization with finer textures.
\section{Network Architecture of DRSC $\mathcal{F}$ }
As mentioned in main manuscript, $\mathcal{F}$ contains a RS correction module and a GS reconstruction module. 
In the following, let us first present the detailed architecture of FlowNet in RS correction module for estimating relative motion map $\bm V_{g\rightarrow t2b}^{(\iota)}$ and $\bm V_{g\rightarrow b2t}^{(\iota)}$ between latent GS images and input RS images. 
Then we elaborate the structure of encoder-decoder in GS reconstruction module for fusing warped images to reconstruct latent GS images $\hat{\bm I}_{g}^{(\iota)}$, where $\iota \in \{t_1,t_m,t_H\}$.

\subsection{Network Architecture of FlowNet in RS Correction Module}
\begin{figure*}[!t]\footnotesize
	\setlength{\abovecaptionskip}{3pt} 
	\setlength{\belowcaptionskip}{0pt}
	\begin{tabular}{cccccc}
		\includegraphics[width=0.999\linewidth]{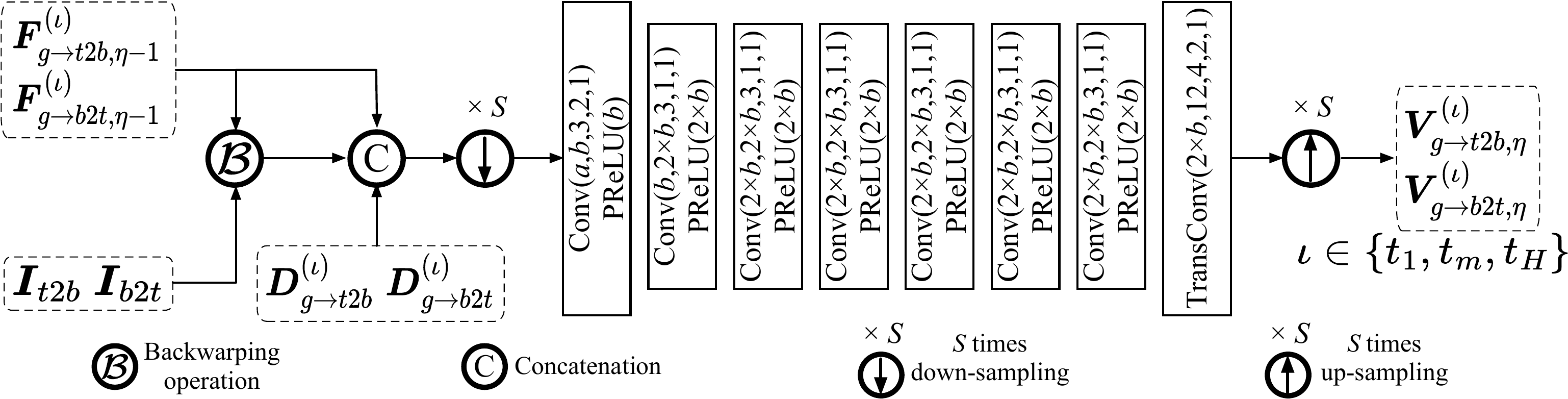}\\
	\end{tabular}
	\caption{
		Illustration of $\eta$-th subnetwork in FlowNet, where $\eta\!\in\!\{1,2,3,4\}$. 
		Before convolution, the warped dual images $\mathcal{B}(\bm I_{t2b};\bm F_{g\rightarrow t2b,\eta-1}^{(\iota)})$ and $\mathcal{B}(\bm I_{b2t};\bm F_{g\rightarrow b2t,\eta-1}^{(\iota)})$, previously estimated dual optical flows $\bm F_{g\rightarrow t2b,\eta-1}^{(\iota)}$ and $\bm F_{g\rightarrow b2t,\eta-1}^{(\iota)}$, and time displacements $\bm D_{g\rightarrow t2b}^{(\iota)}$ and $\bm D_{g\rightarrow b2t}^{(\iota)}$ are concatenated.
		The resolution of concatenation is scaled $S$ times by linear interpolation, where $S=2^{4-\eta}$. Convolution is with the form $\mathrm{Conv}(input\, channel, output\, channel, kernel\, size, stride, padding\, size)$. Transposed convolution is the same formulation.
		While channel dimension $b$ is set as $24\times (5-\eta)$, and channel dimension $a$ is set as $2\times (3+2+1)\times 3$ when $\eta \textgreater 1$. For $\eta=1$, we directly take dual RS images and time displacements as input, hence $a$ is $2\times (3 + 3\times 1)$.
	}
	\label{fig:flownet}
\end{figure*}

Following IFED \cite{zhong2022bringing}, FlowNet totally utilizes 4 subnetworks to iteratively take warped dual images, time displacements and previously estimated optical flows as inputs for estimating current relative motion map between latent GS images and the input RS images. 
These subnetworks share the same structure and are detailed in Fig. \ref{fig:flownet}. 
Specially, the estimated optical flows in $\eta$-th subnetwork ($\eta\!\in\!\{1,2,3,4\}$) are calculated as follows:
\vspace{-1em}
\begin{equation}
	\begin{aligned}
		\bm F_{g\rightarrow t2b,\eta}^{(\iota)}=\sum_{j=1}^{\eta}\bm V_{g\rightarrow t2b,j}^{(\iota)}\odot \bm D_{g\rightarrow t2b}^{(\iota)}, \\
		\bm F_{g\rightarrow b2t,\eta}^{(\iota)}=\sum_{j=1}^{\eta}\bm V_{g\rightarrow b2t,j}^{(\iota)}\odot \bm D_{g\rightarrow b2t}^{(\iota)}, \\
	\end{aligned}
	\vspace{-1em}
\end{equation}
where $\odot$ is an element-wise multiplication. 
The estimated optical flows in $4$-th subnetwork is as the final optical flows $\bm F_{g\rightarrow t2b}^{(\iota)}$ and $\bm F_{g\rightarrow b2t}^{(\iota)}$. 
Note optical flow estimation in $1$-st subnetwork is realized without previously estimated optical flows, \ie, only taking dual RS images and time displacements as input.
%

\subsection{Network Architecture of GS Reconstruction Module}
\begin{figure*}[!t]\footnotesize
	\setlength{\abovecaptionskip}{3pt} 
	\setlength{\belowcaptionskip}{0pt}
	\begin{tabular}{cccccc}
		\includegraphics[width=1\linewidth]{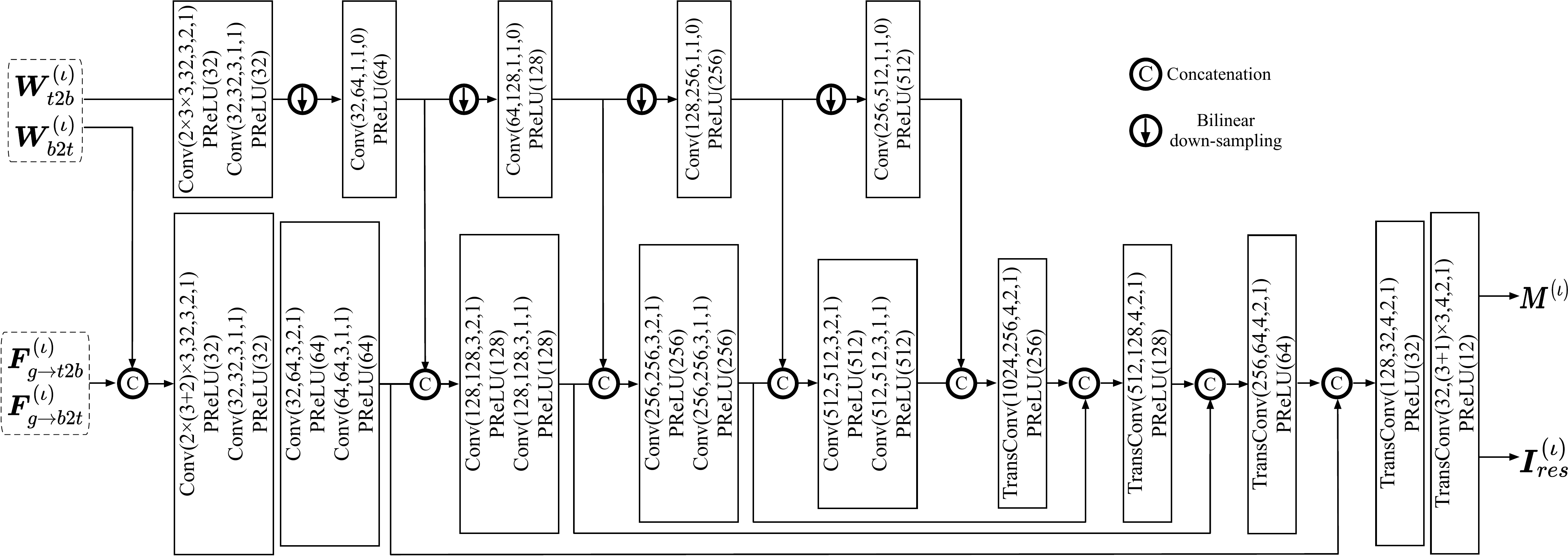}\\
	\end{tabular}
	\hspace{-1em}
	\caption{
		Illustration of encoder-decoder in GS reconstruction module. It can be divided into two parts, \ie, one is the upper part of the figure and the other is the lower part.
		For the upper part, we take only warped dual images $\bm W_{t2b}^{(\iota)}$ and $\bm W_{b2t}^{(\iota)}$ as input for extracting warped features in each level. 
		The warped dual images $\bm W_{t2b}^{(\iota)}$ and $\bm W_{b2t}^{(\iota)}$, and estimated dual optical flows $\bm F_{g\rightarrow t2b}^{(\iota)}$ and $\bm F_{g\rightarrow b2t}^{(\iota)}$ are fed into the lower part. 
		And we concatenate the features of the upper part and the lower part in the $2,3,4,5$ level for feature fusion.
		Finally, the decoder output the fusing mask $\bm M^{(\iota)}$ and residual images $\bm I_{res}^{(\iota)}$ for fusing warped images to reconstruct final GS images.
		Convolution is with the form $\mathrm{Conv}(input\, channel, output\, channel, kernel\, size, stride, padding\, size)$. Transposed convolution is the same formulation.
	}
	\label{fig:Unet}
\end{figure*}

For fusing warped dual images $\bm W_{t2b}^{(\iota)}$ and $\bm W_{b2t}^{(\iota)}$, an encoder-decoder network is adopted as shown in Fig. \ref{fig:Unet}. 
The output fusing masks $\bm M^{(\iota)}$ and residual images $\bm I_{res}^{(\iota)}$ are utilized to reconstruct GS images $\hat{\bm I}_{g}^{(\iota)}$ according to Eq. \eqref{eq:drsc} in main manuscript.

\begin{table}[!t]\footnotesize  
	\centering
	\setlength{\abovecaptionskip}{0pt} 
	\setlength{\belowcaptionskip}{0pt}
	\begin{tabular}{l|c|c}
		\hline
		
		\hline	
		&  NIQE$\downarrow$ / NRQM$\uparrow$ / PI$\downarrow$  &  PSNR$\uparrow$ / SSIM$\uparrow$ / LPIPS$\downarrow$  \\
		
		\hline
		\multicolumn{3}{l}{\textbf{Momentum coefficient $c$ in} $\mathcal{L}_{sd}$:} \\
		
		\hline
		c = 0.9 &  3.433 / 6.905 / 2.975   &  25.362 / 0.8164 / 0.0620  \\
		c = 0.99 & 3.353 / 6.918 / 2.928 &  25.342 / 0.8305 / 0.0638 \\
		c = 0.999 & 3.353 / 6.916 / 2.931   &  27.045 / 0.8618 / 0.0562 \\
		c = 1 (Ours)  &  3.297 / 6.933 / 2.896 &  28.704 / 0.8886 / 0.0546  \\
		
		\hline
		\multicolumn{3}{l}{\textbf{Boundary cropping size $p$ in} $\mathcal{L}_{sd}$:} \\
		
		\hline
		p = 0 (Base) &  3.304 / 6.931 / 2.902 &  28.411 / 0.8848 / 0.0574  \\
		p = 16   &  3.297 / 6.927 / 2.898 &  28.589 / 0.8874 / 0.0545   \\
		p = 32 (Ours) &  3.297 / 6.933 / 2.896 &  28.704 / 0.8886 / 0.0546  \\
		
		p = 64 &  3.308 / 6.939 / 2.898  &  28.706 / 0.8879 / 0.0549  \\
		
		\hline
		\multicolumn{3}{l}{\textbf{Stage number  $N$ in $\mathcal{L}_{sd}$}:} \\

		\hline
		N = 1 (Base)  &  3.304 / 6.931 / 2.902 &  28.411 / 0.8848 / 0.0574    \\
		N = 2 (Ours) &  3.297 / 6.933 / 2.896 &  28.704 / 0.8886 / 0.0546   \\
		N = 3  &  3.305 / 6.940 / 2.895  &  28.761 / 0.8887 / 0.0541   \\
		N = 4 &  3.311 / 6.945 / 2.895   &  28.809 / 0.8890 / 0.0541   \\
		
		%
		
		\hline
		
		\hline
	\end{tabular}
	\caption{Ablation study for momentum coefficient, boundary cropping size and training stage number in self-distillation on RS-GOPRO dataset.}\label{tabel:abl}
\end{table}
\section{More Ablation Study}
For analyzing more details in self-distillation, we implement different momentum update coefficients $c$ for teacher model, different boundary cropping sizes $p$ in boundary cropping operation $\mathcal{C}$, and different training stage numbers $N$ in self-distillation. 

\textit{\textbf{Momentum Update Coefficients.}}
For the update strategy of teacher model (\ie, $\mathcal{F}$ in Stage 1) in self-distillation, we freeze parameters of model in Stage 1 in main manuscript, which can be regarded as a special case (\ie, $c=1$) of momentum update strategy \cite{he2020momentum}.
Formally, denoting the parameters of $\mathcal{F}$ in Stage 1 as $\bm{\Theta}_{n=1}$ and those of $\mathcal{F}$ in Stage 2 as $\bm{\Theta}_{n=2}$, we update $\bm{\Theta}_{n=1}$ by:
\begin{equation}
	\bm{\Theta}_{n=1} = c\bm{\Theta}_{n=1} + (1-c)\bm{\Theta}_{n=2}
\end{equation}
where $c\!\in\![0,1]$ is a momentum coefficient.
Only the parameters $\bm{\Theta}_{n=2}$ are updated by back-propagation through Eq. (15) in main manuscript.
And we implement different momentum update coefficients $c\!\in\!\{0.9,0.99,0.999,1\}$ in Table \ref{tabel:abl}. One can see that model in Stage 1 with frozen parameters can achieve the best performance.
This is because the role of model in Stage 1 is to ensure that the center region of the results is stably provided as pseudo GS images for training model in Stage 2. Hence, we set $c=1$ in main manuscript for better performance.

\textit{\textbf{Boundary Cropping Size.}}
For boundary cropping size $p$ in boundary cropping operation $\mathcal{C}$, we explore how much boundary cropping size can achieve the purpose of mitigating boundary artifacts.
We implement different sizes $p\!\in\!\{16,32,64\}$ and we find $p \geq 32$ can achieve a better performance in Table \ref{tabel:abl}. 
Note that $p=0$ indicates our model training with only $\mathcal{L}_{self}$. And we find no significant improvement in terms of IQA values when $p=64$. 
Finally, we set $p=32$ in our training process.

\textit{\textbf{Number of Training Stages for Self-distillation.}}
For training stage $N$ in self-distillation, we set different $N$ in Table \ref{tabel:abl}. We can observe the improvement is small or fluctuates when $N \geq 2$. Note that extra training stages need more training time (\ie, 115K iterations for each extra stage). 
And we find $N=2$ is sufficient to mitigate boundary artifacts in Fig. \ref{fig:boundary} in main manuscript. Considering the balance of training time and performance, we finally set $N=2$.
\begin{figure}[!t]\footnotesize
	\hspace{-3em}
	\begin{tabular}{cccccc}
		\includegraphics[width=1.1\linewidth]{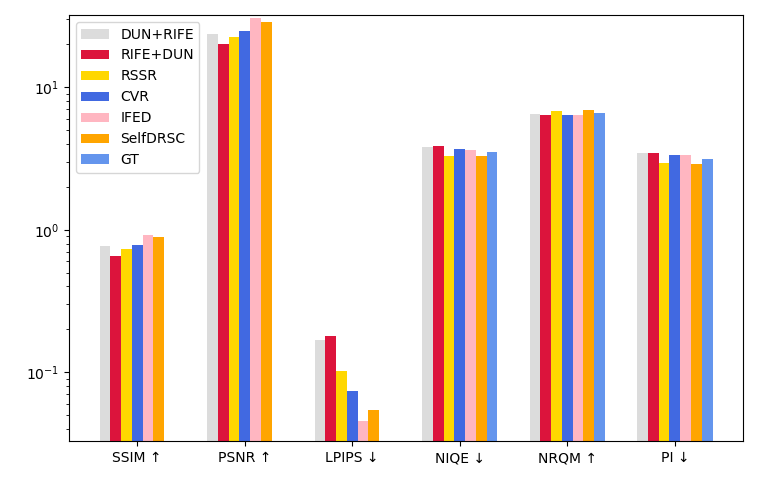}\\
	\end{tabular}
	\vspace{-1em}
	\caption{Overall quantitative justification on RS-GOPRO dataset \cite{zhong2022bringing}.
	}\label{fig:bar}
\end{figure}
\section{More Results on Synthetic Datasets and Real-world Data}
Due to interpolated frames in ground-truth (GT), FR-IQA metrics are not so trustworthy for this task.  Commonly, NR-IQA metrics are not very robust. 
So these quantitative metrics are all adopted as reference for evaluation, and we present the overall quantitative justification of our method using a bar chart in Fig.~\ref{fig:bar}.
In addition, we compare SelfDRSC with IFED which is a top-tier method for dual reversed RS correction.
As mentioned in main manuscript, interpolated frames by VFI method may appear in 9 ground-truth GS images, which leads to poor quality of ground-truth (GT).
We give more results to testify that our SelfDRSC can generate well corrected GS images with higher framerate and finer textures than GT and fully supervised method IFED.
%

{\small
	\bibliographystyle{ieee_fullname}
	\bibliography{egbib}

\begin{thebibliography}{10}\itemsep=-1pt

\bibitem{albl2020two}
Cenek Albl, Zuzana Kukelova, Viktor Larsson, Michal Polic, Tomas Pajdla, and
  Konrad Schindler.
\newblock From two rolling shutters to one global shutter.
\newblock In {\em CVPR}, pages 2505--2513, 2020.

\bibitem{albl2015r6p}
Cenek Albl, Zuzana Kukelova, and Tomas Pajdla.
\newblock R6p-rolling shutter absolute camera pose.
\newblock In {\em CVPR}, pages 2292--2300, 2015.

\bibitem{bai2022self}
Haoran Bai and Jinshan Pan.
\newblock Self-supervised deep blind video super-resolution.
\newblock {\em {ArXiv} preprint arXiv:2201.07422}, 2022.

\bibitem{blau20182018}
Yochai Blau, Roey Mechrez, Radu Timofte, Tomer Michaeli, and Lihi Zelnik-Manor.
\newblock The 2018 pirm challenge on perceptual image super-resolution.
\newblock In {\em ECCV Workshops}, pages 0--0, 2018.

\bibitem{chen2018reblur2deblur}
Huaijin Chen, Jinwei Gu, Orazio Gallo, Ming-Yu Liu, Ashok Veeraraghavan, and
  Jan Kautz.
\newblock Reblur2deblur: Deblurring videos via self-supervised learning.
\newblock In {\em ICCP}, pages 1--9, 2018.

\bibitem{Fan_2021_ICCV}
Bin Fan and Yuchao Dai.
\newblock Inverting a rolling shutter camera: {B}ring rolling shutter images to
  high framerate global shutter video.
\newblock In {\em ICCV}, pages 4228--4237, 2021.

\bibitem{Fan_2022_CVPR}
Bin Fan, Yuchao Dai, Zhiyuan Zhang, Qi Liu, and Mingyi He.
\newblock Context-aware video reconstruction for rolling shutter cameras.
\newblock In {\em CVPR}, pages 17572--17582, 2022.

\bibitem{grill2020bootstrap}
Jean-Bastien Grill, Florian Strub, Florent Altch{\'e}, Corentin Tallec, Pierre
  Richemond, Elena Buchatskaya, Carl Doersch, Bernardo Avila~Pires, Zhaohan
  Guo, Mohammad Gheshlaghi~Azar, et~al.
\newblock Bootstrap your own latent-a new approach to self-supervised learning.
\newblock {\em NeurIPS}, 33:21271--21284, 2020.

\bibitem{he2020momentum}
Kaiming He, Haoqi Fan, Yuxin Wu, Saining Xie, and Ross Girshick.
\newblock Momentum contrast for unsupervised visual representation learning.
\newblock In {\em CVPR}, pages 9729--9738, 2020.

\bibitem{huang2022rife}
Zhewei Huang, Tianyuan Zhang, Wen Heng, Boxin Shi, and Shuchang Zhou.
\newblock Real-time intermediate flow estimation for video frame interpolation.
\newblock In {\em ECCV}, pages 1--16, 2022.

\bibitem{jiang2018super}
Huaizu Jiang, Deqing Sun, Varun Jampani, Ming-Hsuan Yang, Erik Learned-Miller,
  and Jan Kautz.
\newblock Super {S}lomo: {H}igh quality estimation of multiple intermediate
  frames for video interpolation.
\newblock In {\em CVPR}, pages 9000--9008, 2018.

\bibitem{johnson2016perceptual}
Justin Johnson, Alexandre Alahi, and Li Fei-Fei.
\newblock Perceptual losses for real-time style transfer and super-resolution.
\newblock In {\em ECCV}, pages 694--711, 2016.

\bibitem{lai2018fast}
Wei-Sheng Lai, Jia-Bin Huang, Narendra Ahuja, and Ming-Hsuan Yang.
\newblock Fast and accurate image super-resolution with deep laplacian pyramid
  networks.
\newblock {\em IEEE TPAMI}, 41(11):2599--2613, 2018.

\bibitem{lao2020rolling}
Yizhen Lao and Omar Ait-Aider.
\newblock Rolling shutter homography and its applications.
\newblock {\em IEEE TPAMI}, 43(8):2780--2793, 2020.

\bibitem{litwiller2001ccd}
Dave Litwiller.
\newblock {CCD} vs. {CMOS}.
\newblock {\em Photonics spectra}, 35(1):154--158, 2001.

\bibitem{liu2020deep}
Peidong Liu, Zhaopeng Cui, Viktor Larsson, and Marc Pollefeys.
\newblock Deep shutter unrolling network.
\newblock In {\em CVPR}, pages 5941--5949, 2020.

\bibitem{liu2020self}
Peidong Liu, Joel Janai, Marc Pollefeys, Torsten Sattler, and Andreas Geiger.
\newblock Self-supervised linear motion deblurring.
\newblock {\em IEEE Robotics and Automation Letters}, 5(2):2475--2482, 2020.

\bibitem{loshchilov2016sgdr}
Ilya Loshchilov and Frank Hutter.
\newblock {SGDR}: Stochastic gradient descent with warm restarts.
\newblock {\em {ArXiv} preprint arXiv:1608.03983}, 2016.

\bibitem{loshchilov2017decoupled}
Ilya Loshchilov and Frank Hutter.
\newblock Decoupled weight decay regularization.
\newblock {\em {ArXiv} preprint arXiv:1711.05101}, 2017.

\bibitem{ma2017learning}
Chao Ma, Chih-Yuan Yang, Xiaokang Yang, and Ming-Hsuan Yang.
\newblock Learning a no-reference quality metric for single-image
  super-resolution.
\newblock {\em CVIU}, 158:1--16, 2017.

\bibitem{meingast2005geometric}
Marci Meingast, Christopher Geyer, and Shankar Sastry.
\newblock Geometric models of rolling-shutter cameras.
\newblock {\em {ArXiv} preprint cs/0503076}, 2005.

\bibitem{mittal2012making}
Anish Mittal, Rajiv Soundararajan, and Alan~C Bovik.
\newblock Making a “completely blind” image quality analyzer.
\newblock {\em IEEE Signal processing letters}, 20(3):209--212, 2012.

\bibitem{niklaus2020softmax}
Simon Niklaus and Feng Liu.
\newblock Softmax splatting for video frame interpolation.
\newblock In {\em CVPR}, pages 5437--5446, 2020.

\bibitem{ren2020neural}
Dongwei Ren, Kai Zhang, Qilong Wang, Qinghua Hu, and Wangmeng Zuo.
\newblock Neural blind deconvolution using deep priors.
\newblock In {\em CVPR}, pages 3341--3350, 2020.

\bibitem{sim2021xvfi}
Hyeonjun Sim, Jihyong Oh, and Munchurl Kim.
\newblock {XVFI}: {E}xtreme video frame interpolation.
\newblock In {\em ICCV}, pages 14489--14498, 2021.

\bibitem{sun2018pwc}
Deqing Sun, Xiaodong Yang, Ming-Yu Liu, and Jan Kautz.
\newblock {PWC-Net}: {CNN}s for optical flow using pyramid, warping, and cost
  volume.
\newblock In {\em CVPR}, pages 8934--8943, 2018.

\bibitem{wang2004image}
Zhou Wang, Alan~C Bovik, Hamid~R Sheikh, and Eero~P Simoncelli.
\newblock Image quality assessment: {F}rom error visibility to structural
  similarity.
\newblock {\em IEEE TIP}, 13(4):600--612, 2004.

\bibitem{xu2022gmflow}
Haofei Xu, Jing Zhang, Jianfei Cai, Hamid Rezatofighi, and Dacheng Tao.
\newblock {GMFlow}: Learning optical flow via global matching.
\newblock In {\em CVPR}, pages 8121--8130, 2022.

\bibitem{xu2019quadratic}
Xiangyu Xu, Li Siyao, Wenxiu Sun, Qian Yin, and Ming-Hsuan Yang.
\newblock Quadratic video interpolation.
\newblock {\em NeurIPS}, 32:1--10, 2019.

\bibitem{zhang2018unreasonable}
Richard Zhang, Phillip Isola, Alexei~A Efros, Eli Shechtman, and Oliver Wang.
\newblock The unreasonable effectiveness of deep features as a perceptual
  metric.
\newblock In {\em CVPR}, pages 586--595, 2018.

\bibitem{zhong2022bringing}
Zhihang Zhong, Mingdeng Cao, Xiao Sun, Zhirong Wu, Zhongyi Zhou, Yinqiang
  Zheng, Stephen Lin, and Imari Sato.
\newblock Bringing rolling shutter images alive with dual reversed distortion.
\newblock In {\em ECCV}, pages 233--249, 2022.

\bibitem{zhu2017unpaired}
Jun-Yan Zhu, Taesung Park, Phillip Isola, and Alexei~A Efros.
\newblock Unpaired image-to-image translation using cycle-consistent
  adversarial networks.
\newblock In {\em ICCV}, pages 2223--2232, 2017.

\bibitem{zhuang2019learning}
Bingbing Zhuang, Quoc-Huy Tran, Pan Ji, Loong-Fah Cheong, and Manmohan
  Chandraker.
\newblock Learning structure-and-motion-aware rolling shutter correction.
\newblock In {\em CVPR}, pages 4551--4560, 2019.

\end{thebibliography}
}

\begin{figure*}[!t]\footnotesize
	\centering
	\setlength{\abovecaptionskip}{3pt} 
	\setlength{\belowcaptionskip}{0pt}
	\begin{tabular}{cccccc}
		\includegraphics[width=0.9\linewidth]{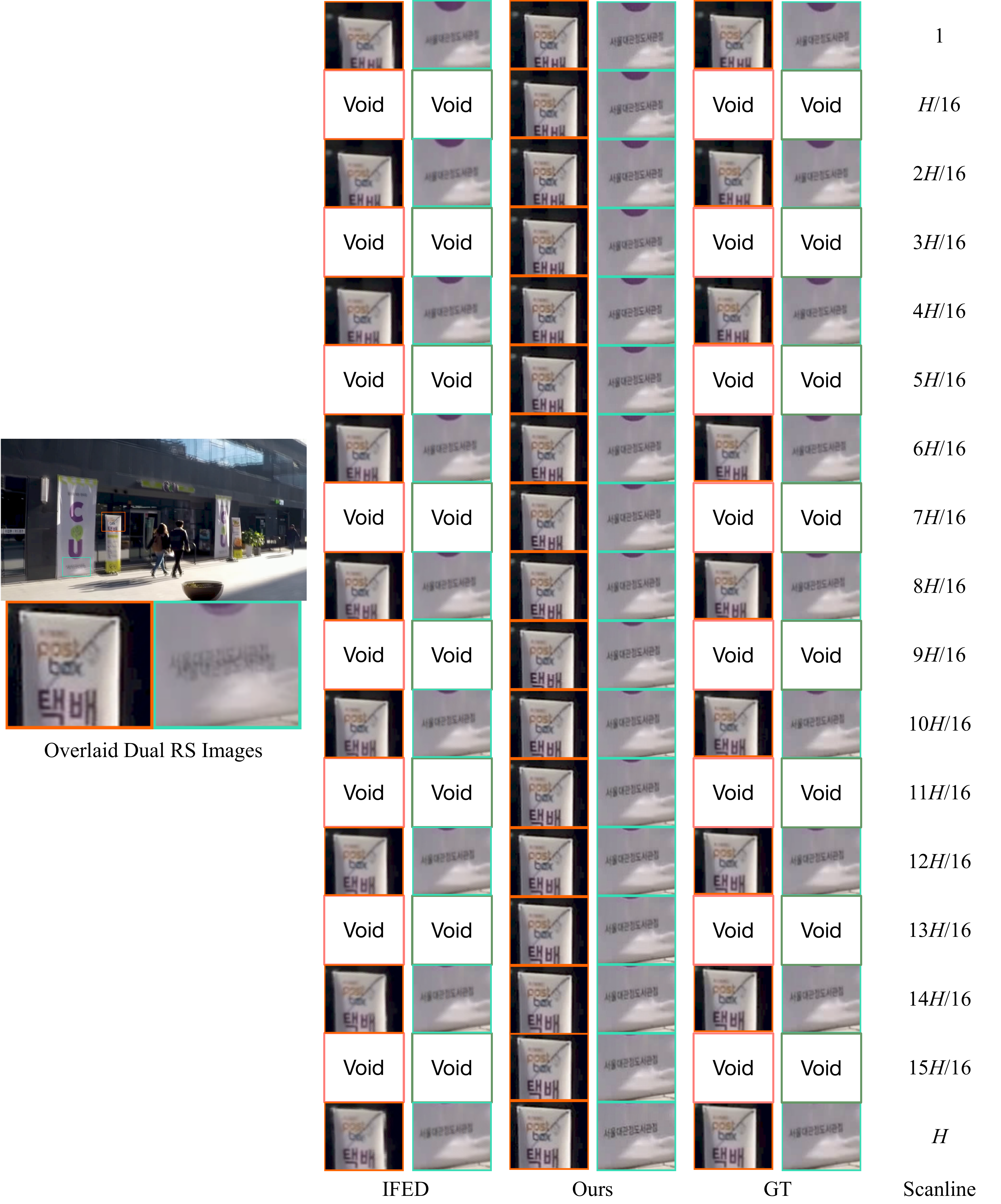}\\
	\end{tabular}
	\hspace{-1em}
\end{figure*}
\clearpage

\begin{figure*}[!t]\footnotesize
	\centering
	\setlength{\abovecaptionskip}{3pt} 
	\setlength{\belowcaptionskip}{0pt}
	\begin{tabular}{cccccc}
		\includegraphics[width=0.85\linewidth]{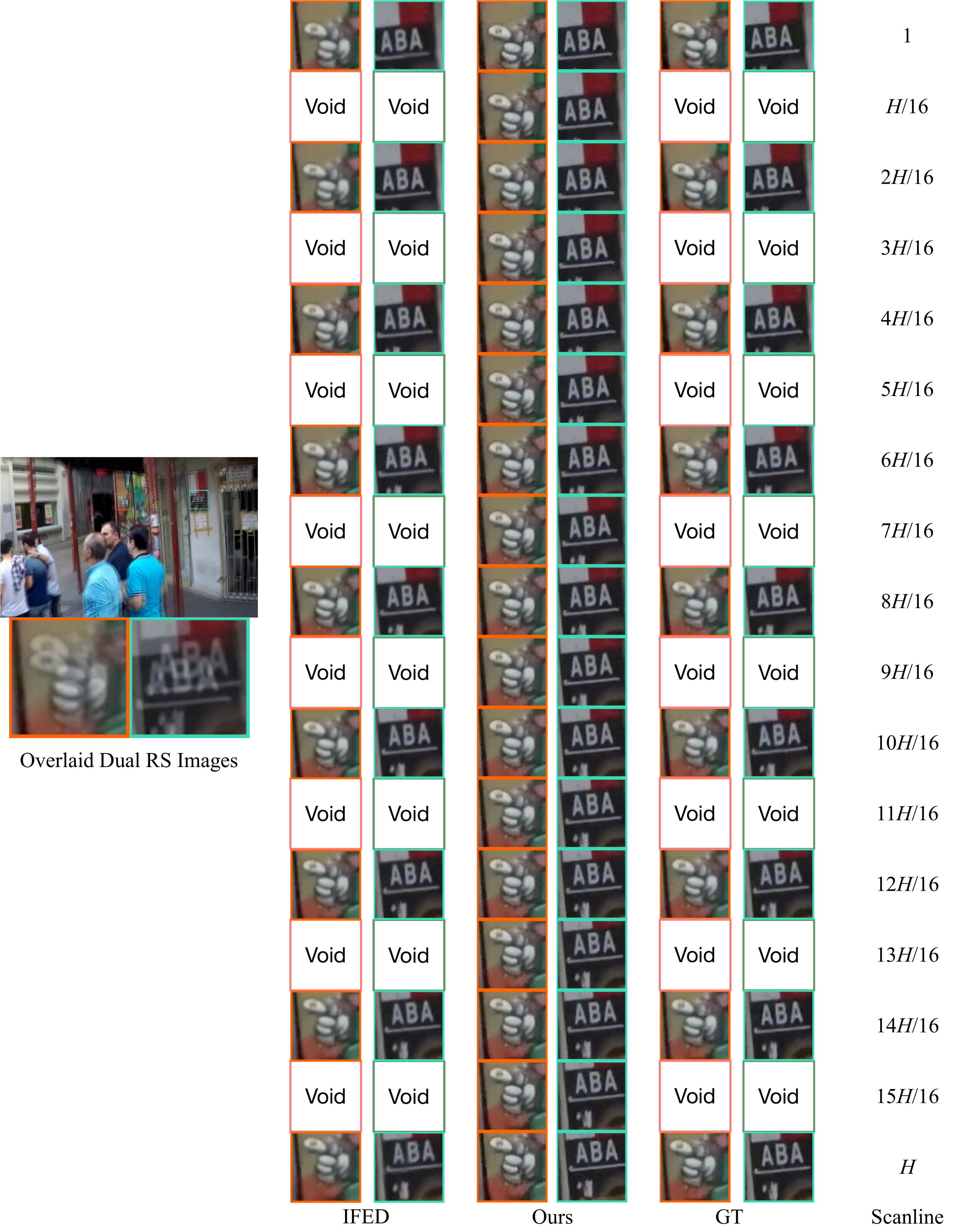}\\
	\end{tabular}
	\hspace{-1em}
\end{figure*}
\clearpage

\begin{figure*}[!t]\footnotesize
	\centering
	\setlength{\abovecaptionskip}{3pt} 
	\setlength{\belowcaptionskip}{0pt}
	\begin{tabular}{cccccc}
		\includegraphics[width=1\linewidth]{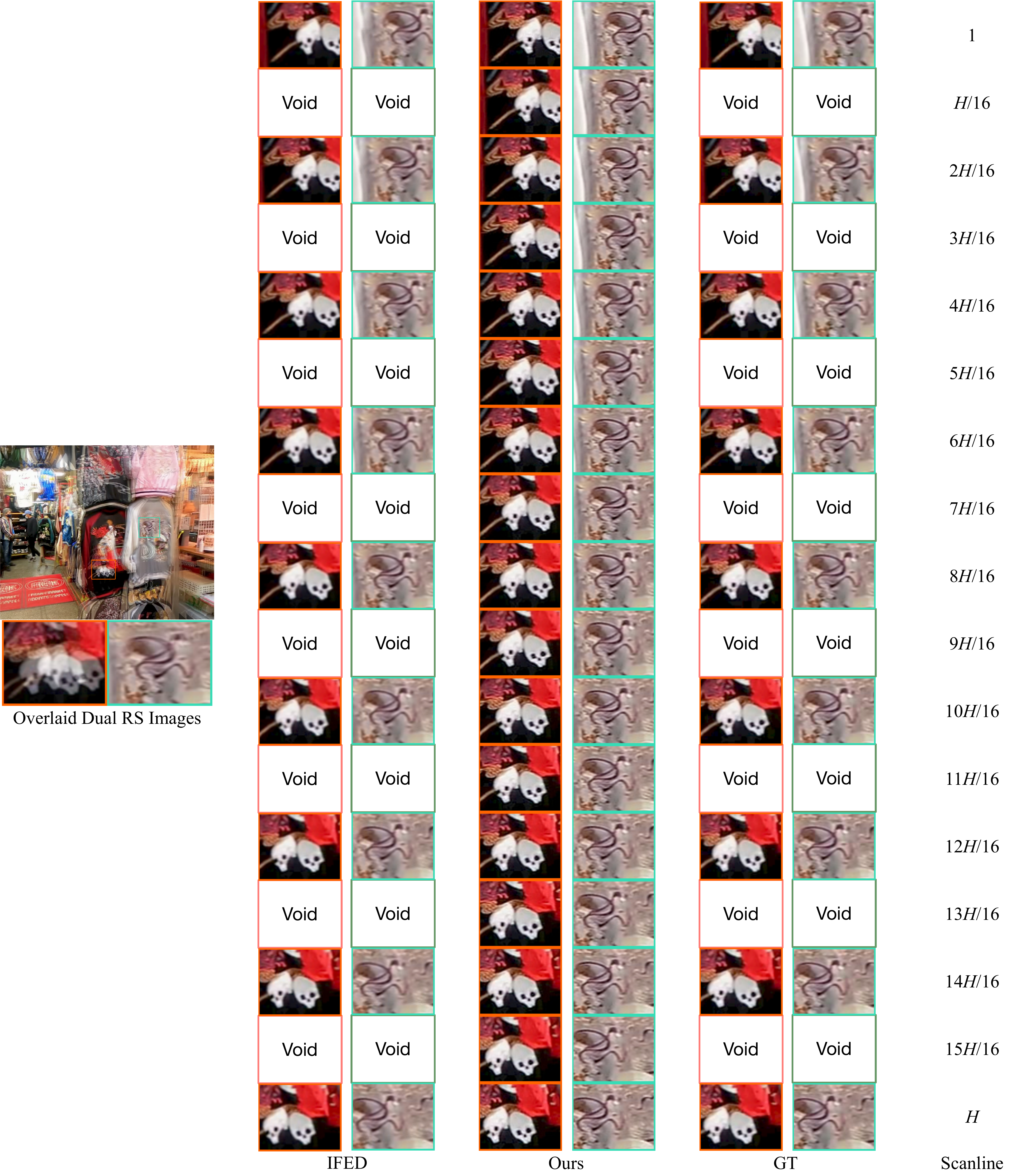}\\
	\end{tabular}
	\hspace{-1em}
\end{figure*}
\clearpage

\begin{figure*}[!t]\footnotesize
	\centering
	\setlength{\abovecaptionskip}{3pt} 
	\setlength{\belowcaptionskip}{0pt}
	\begin{tabular}{cccccc}
		\includegraphics[width=1\linewidth]{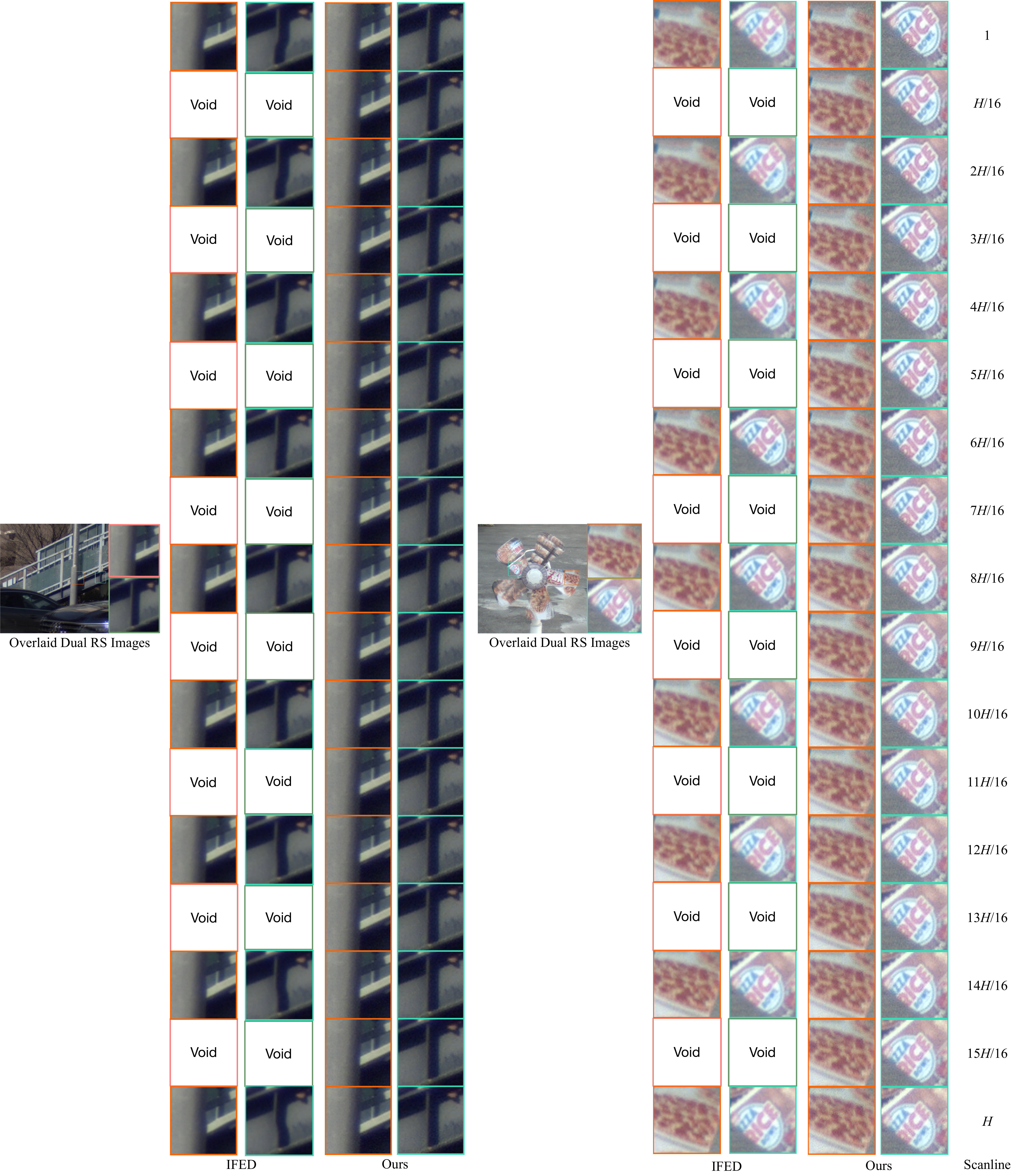}\\
	\end{tabular}
	\hspace{-1em}
\end{figure*}

\end{document}